%% file: main.tex
\documentclass[10pt,twocolumn,letterpaper]{article}

 \usepackage[pagenumbers]{cvpr} 

\input{preamble}

\definecolor{cvprblue}{rgb}{0.21,0.49,0.74}
\usepackage[pagebackref,breaklinks,colorlinks,allcolors=cvprblue]{hyperref}

\def\confName{CVPR}
\def\confYear{2027}

\title{KeyDeck: Multi-Paper-to-Slide Generation \\via Cross-Paper Synthesis and Reference-Guided Personalization}

\newcommand{\authorsep}{\hspace{1.6em}}

\author{
Tarik Can Ozden \authorsep Sachidanand VS \authorsep Ram Prakash Reddy Motati \\
Ozgur Kara \authorsep Dilek Hakkani-Tür \authorsep Junho Kim$^\dagger$ \authorsep James M. Rehg$^\dagger$ \\[0.6em]
University of Illinois Urbana-Champaign
}

\newcommand{\methodname}{KeyDeck\xspace}

\begin{document}
\maketitle

\begin{strip}
    \centering
    \vspace{-8mm}
    \includegraphics[width=\textwidth]{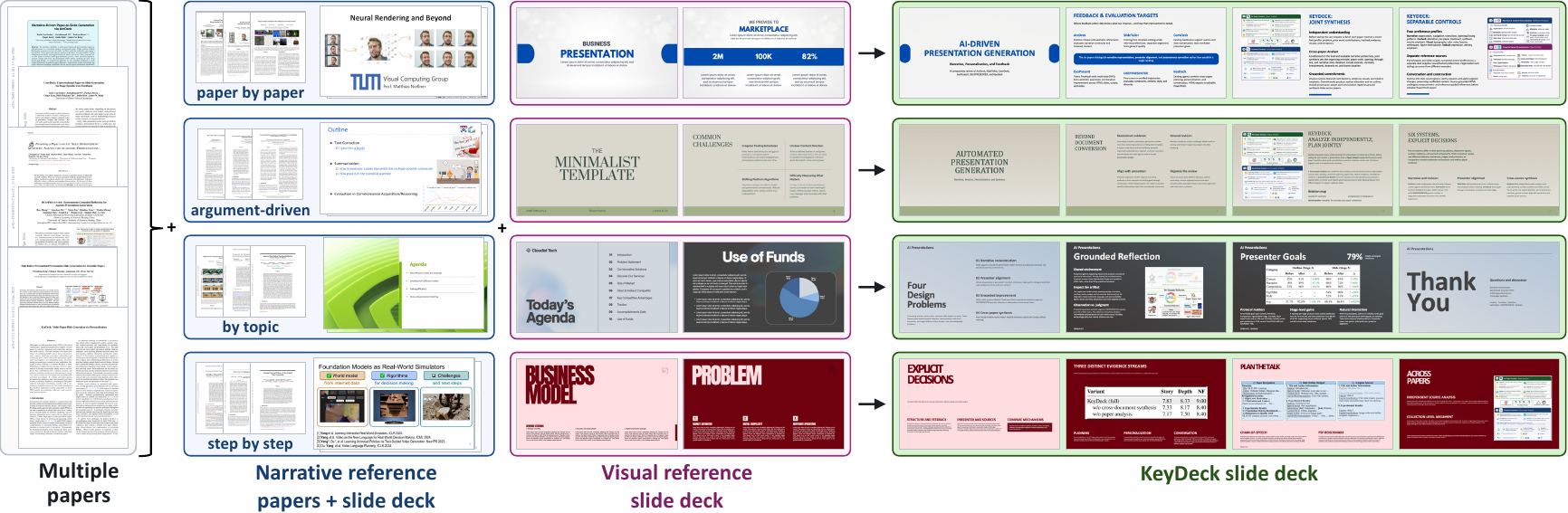}
    \vspace{-5mm}
    \captionsetup{hypcap=false}
    \captionof{figure}{\textbf{Overview of KeyDeck.} Given multiple source papers, KeyDeck generates a coherent slide deck personalized by narrative references (papers and their corresponding slide decks) and a separate visual reference slide deck. The narrative references guide content organization and presentation structure, while the visual reference guides the generated deck's visual and textual style.}
    \label{fig:teaser}
    \captionsetup{hypcap=true}
\end{strip}

\begin{abstract}
Multi-paper-to-slide generation (multi-PTSG) is the task of constructing a unified presentation from multiple research papers by selecting, organizing, and synthesizing information across sources. This task introduces two main challenges: (i) combining multiple sources into a coherent narrative requires identifying cross-paper relationships and synthesizing their content, and (ii) the desired organization and emphasis varies substantially across presenters, making personalization essential for user satisfaction. Motivated by these challenges, we propose \methodname, the first multi-PTSG framework that integrates per-paper and cross-paper analysis to identify relationships among sources and synthesize their contributions into a coherent narrative, and reference-guided personalization to infer presenters' narrative, content, textual, and visual preferences and adapt slide generation accordingly. To evaluate generation quality, personalization adherence, and conversational goal satisfaction, we introduce \textit{KeyBench}, a benchmark of 100 expert-authored multi-paper presentations, together with a comprehensive evaluation framework. Our results demonstrate that \methodname outperforms existing approaches in multi-PTSG quality, personalization adherence, and conversational goal satisfaction. More details are in our \textcolor{blue}{\href{https://keydeck-slide.github.io}{project webpage}}.

\end{abstract}

\section{Introduction}

\begingroup
\renewcommand{\thefootnote}{}
\footnotetext{$^\dagger$Corresponding author}
\endgroup
\setcounter{footnote}{0}

\begin{figure*}[t]
    \centering
    \includegraphics[width=0.8\textwidth]{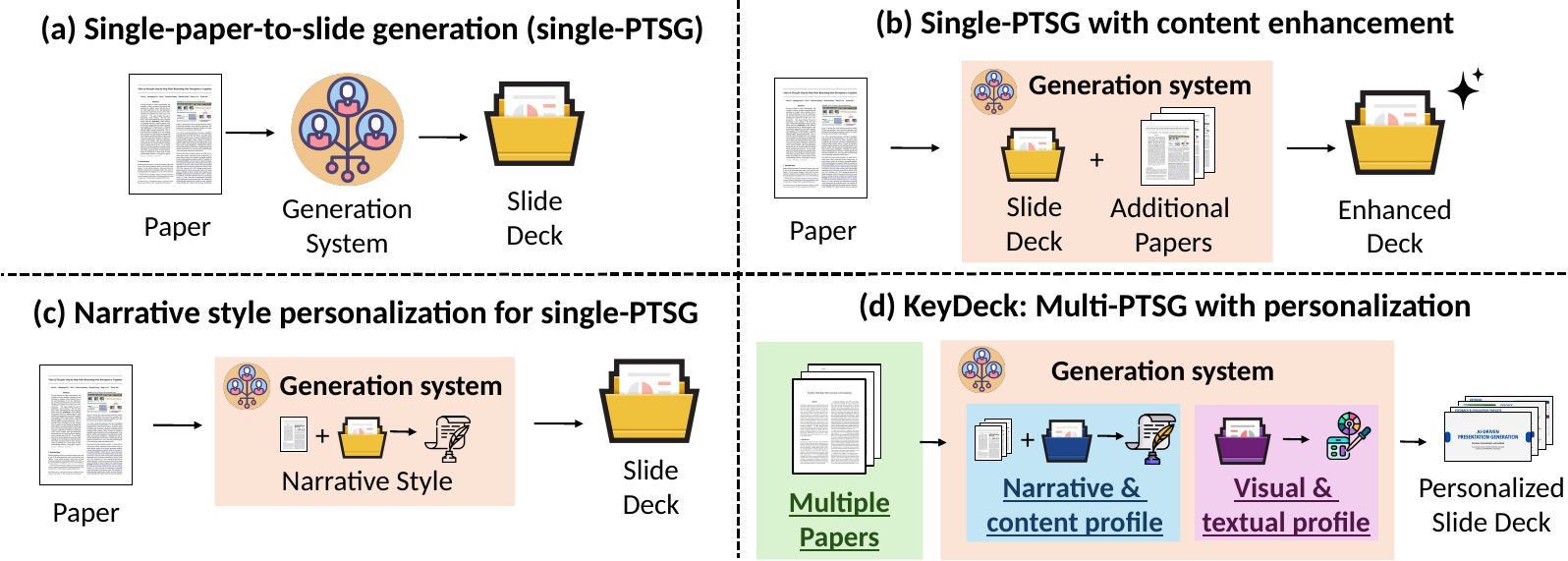}
    \vspace{-3mm}
    \caption{\textbf{Comparison of paper-to-slide generation paradigms.}
    Previous work focuses on (a) single-paper-to-slide generation (single-PTSG), (b) single-PTSG with supplementary content from additional papers,
    (c) single-PTSG with narrative-style personalization, while (d) our framework jointly synthesizes
\textcolor[RGB]{50,82,32}{\textbf{multiple papers}}
with \textcolor[RGB]{33,61,97}{\textbf{narrative, content}},
\textcolor[RGB]{73,26,72}{\textbf{visual and textual}} personalization.}
    \label{fig:comparison}
\end{figure*}

Presentations are a primary medium of communication that typically synthesize information from multiple sources, particularly in research tutorials, surveys, and technical talks. We define multi-paper-to-slide generation (multi-PTSG) as the task of producing a unified slide deck from a collection of research papers by selecting, organizing, and synthesizing information across sources into a coherent presentation. Unlike single-paper-to-slide generation (single-PTSG), which focuses on selecting and organizing content from a single source~\cite{liang2025slidegen, ozden2026narrative, ozden2026convdeck, zheng2026deeppresenter}, multi-PTSG requires reasoning across sources to identify relationships, reconcile complementary or contrasting findings, and synthesize information at varying levels of abstraction. Consequently, the resulting presentation must communicate these cross-source connections through a coherent narrative rather than simply concatenating summaries of individual papers.

An additional challenge in multi-PTSG is personalization, which refers to adapting the content, narrative structure, textual expression, and visual design of a presentation to the user's goals and preferences~\cite{zeng2026slidetailor}. The same collection of source papers can support multiple valid presentations, each involving different decisions about content selection, emphasis, abstraction, and narrative organization~\cite{ozden2026narrative}. For instance, a presentation may organize research papers chronologically to illustrate the evolution of a field, compare their methodological differences, or synthesize their findings around shared research themes. Beyond these content and narrative choices, users may also have distinct preferences regarding textual density, visual composition, and slide layout. Such preferences are not always explicitly specified and may instead be reflected in previously authored presentations. Consequently, presentation quality cannot be determined solely by the coverage and factual accuracy of the source material; it also depends on how effectively the generated presentation reflects the intended communicative goals and preferences of the user~\cite{ozden2026convdeck}.

Despite recent advances in automated slide generation, existing approaches predominantly focus on individual source papers~\cite{zheng2025pptagent, liang2025slidegen, ozden2026narrative, ozden2026convdeck, zheng2026deeppresenter}, while support for multiple papers is often limited to supplementary content incorporation or post-hoc refinement rather than joint planning from the outset~\cite{ozden2026convdeck,liu2025presenting}. Similarly, personalization mechanisms have largely relied on explicit user instructions~\cite{ozden2026narrative, zheng2026deeppresenter} or interactive feedback~\cite{ozden2026convdeck}, with limited consideration of preferences implicitly reflected in previously authored presentations~\cite{zeng2026slidetailor}. These limitations highlight the need for a unified framework that performs structured cross-paper synthesis while incorporating user-specific preferences throughout the generation process. In particular, reference presentations provide valuable guidance for identifying how users select, prioritize, and organize source information, as well as their preferred textual and visual characteristics.

To address these challenges, we propose \textit{\methodname}, the first personalization-aware comprehensive framework for multi-PTSG (Fig.~\ref{fig:comparison}). \methodname first analyzes individual source papers and jointly synthesizes their content to identify cross-paper relationships, shared themes, and complementary or contrasting findings. Based on this analysis, it constructs a coherent presentation narrative that integrates information across sources rather than presenting them independently. To support personalization, we leverage a reference-based mechanism that captures user preferences along four dimensions: narrative, content, visual, and textual. By analyzing reference presentations and their associated source papers, \methodname adapts the organization, content selection and emphasis, visual design, and textual expression of generated slides to reflect the user's presentation preferences. Additionally, the framework supports conversational refinement so users can iteratively adjust the generated deck to match their objectives. To evaluate these capabilities, we introduce \textit{KeyBench}, a benchmark of 100 expert-authored multi-paper presentations paired with their source papers, along with evaluation protocols for multi-paper generation quality, personalization adherence, and conversational goal satisfaction. Our results demonstrate superior performance in multi-paper generation quality, personalization, and conversational goal satisfaction compared to prior methods.

\begin{figure*}[t]
    \centering
    \includegraphics[width=0.87\textwidth]{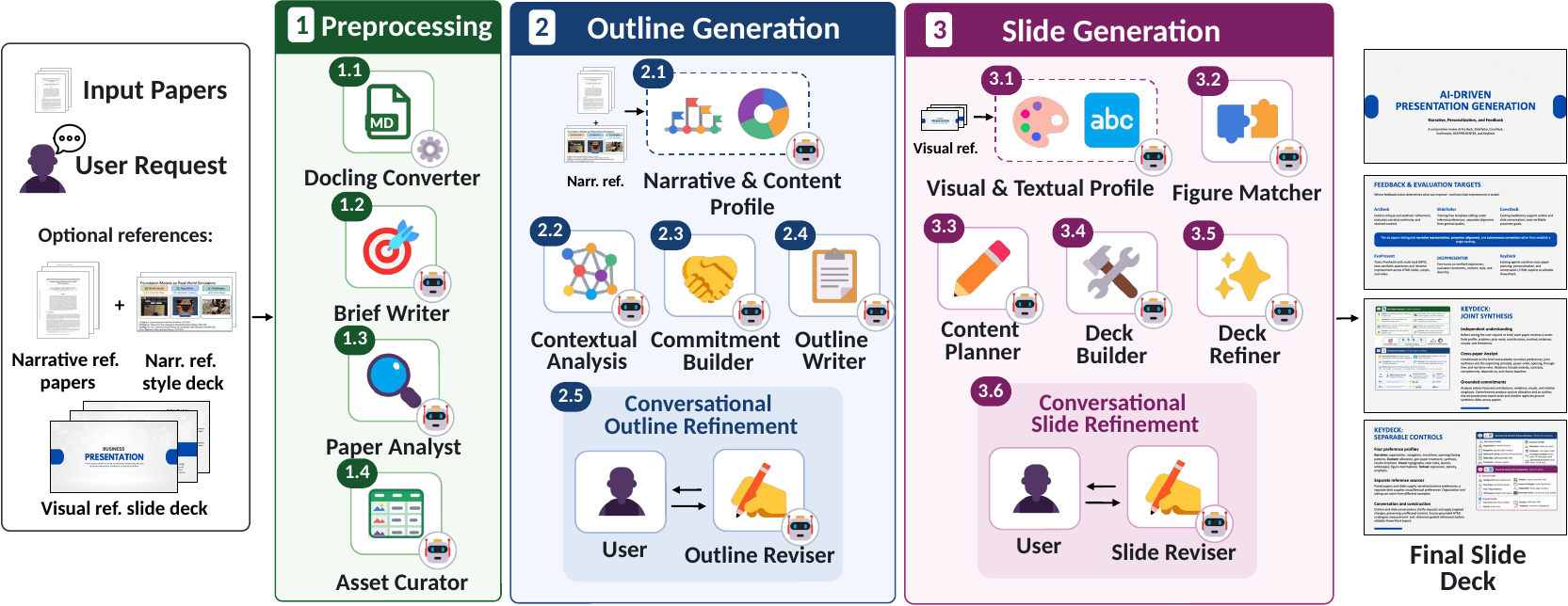}
    \vspace{-3mm}
    \caption{\textbf{Overview of the proposed multi-paper-to-slide generation framework.  }
    (1) \textbf{Preprocessing:} Source papers are converted into Markdown, a presentation brief is derived from the user request, and per-paper analyses, metadata, and visual assets are extracted.
    (2) \textbf{Outline Generation:} Per-paper analyses are integrated through cross-paper contextual analysis to identify relationships and synthesize information. Narrative and content personalization guide narrative planning and slide outline generation.
    (3) \textbf{Slide Generation:} Figures and tables are matched to slides, and the slide deck is constructed using visual and textual personalization, followed by aesthetic refinement to produce the final editable presentation.} 
    \label{fig:pipeline}
\end{figure*}

In summary, our main contributions are as follows:
\begin{itemize}
    \item We introduce \textit{\methodname}, a framework for multi-paper-to-slide generation (multi-PTSG) that integrates structured per-paper analysis and cross-paper synthesis to construct coherent presentation narratives.

    \item We propose a reference-based personalization approach that captures narrative, content, visual, and textual preferences from prior presentations to guide presentation generation.

    \item We introduce \textit{KeyBench}, a benchmark comprising 100 expert-authored multi-paper presentations paired with their corresponding source papers, along with evaluation protocols for multi-PTSG quality, personalization adherence, and conversational goal satisfaction.
\end{itemize}

\section{Related Work}
\paragraph{Slide Generation.}
Early approaches to paper-to-slide generation focused on content selection, summarization, and layout prediction~\cite{sun2021d2s,fu2022doc2ppt}. More recent systems employ LLM-based planning and multi-agent orchestration for slide generation~\cite{cachola2024kctv,bandyopadhyay2024docpress,zheng2025pptagent,ge2025autopresent}. Related works also explore alternative formats for presenting papers, such as posters and videos~\cite{pang2025paper2poster,zhu2025papervideo,yu2026paperx, zhang2025postergen, Inadumi_2026_SciPostGen, shi2026apexacademicposterediting}, while other approaches incorporate iterative refinement and narrative planning~\cite{liang2025slidegen,ozden2026narrative}. However, these approaches predominantly focus on single-PTSG. ConvDeck~\cite{ozden2026convdeck} and EvoPresent~\cite{liu2025presenting} can incorporate additional papers during later refinement stages for content enhancement, but their generation pipelines remain centered on a primary source rather than jointly planning across multiple papers from the outset. DeepPresenter~\cite{zheng2026deeppresenter} employs flexible agentic orchestration that can operate over multiple sources, but does not explicitly capture structured cross-paper narrative planning. Consequently, existing methods offer limited support for multi-PTSG, making their direct application to multiple papers prone to fragmented narratives, redundant content, and insufficient synthesis of cross-paper relationships. Existing evaluations similarly emphasize single-paper settings, with limited consideration of cross-paper synthesis and narrative coherence.

\paragraph{Cross-Paper Synthesis.}
Early approaches to multi-paper synthesis focused on aggregating and organizing information across sources~\cite{lu2020multi,xiao2022primera}. More recent systems employ LLM-based planning, multi-agent collaboration, and iterative refinement for scientific literature analysis, synthesis, and survey or manuscript generation~\cite{wang2024autosurvey,chao2025llm,go2026lira,zheng2026scilens,song2026paperorchestra}. While these approaches primarily focus on textual analysis and generation, we similarly leverage LLM-based cross-paper reasoning to identify relationships among source papers and synthesize their contributions, but extend this process to slide generation by organizing the synthesized information into a coherent presentation narrative with appropriate content allocation and visual evidence.

\paragraph{Slide Personalization.} Prior work has explored several forms of personalization in slide generation. ArcDeck~\cite{ozden2026narrative} conditions generation on user-specified audience and presentation duration, while ConvDeck~\cite{ozden2026convdeck} extends this setting with free-form user feedback through interactive refinement. 
However, these approaches rely primarily on explicit user inputs and do not infer presentation preferences from previous slide decks. SlideTailor~\cite{zeng2026slidetailor} extracts narrative preferences from a paper-slide pair, but it does not model visual or textual preferences from the reference, and instead relies on a template PPTX deck for visual styling.
We extend this direction by extracting complementary preference profiles from reference paper(s)-slide pairs: narrative and content preferences that capture how source material is selected and organized, and a separate reference slide deck for visual and textual preferences that capture how the presentation is styled. 

\section{KeyDeck: Multi-Paper-to-Slide Generation}
\label{sec:method}

\methodname{} takes as input one or more source PDFs $\mathcal{P}=\{P_1,\ldots,P_N\}$, an optional natural-language request $Q$, and optional personalization references. The \emph{narrative reference} consists of papers and the slide deck presenting them, while the \emph{visual reference} is a separate slide deck. These references allow the organization and content emphasis of one presentation to be combined with the visual and textual style of another (Fig.~\ref{fig:personalization_profiles}). As shown in Fig.~\ref{fig:pipeline}, the framework consists of three stages: (1) preprocessing, (2) outline generation, and (3) slide generation. Users can provide feedback on both the outline and the generated slides. The final output is an editable PowerPoint presentation.

\subsection{Preprocessing}
\label{sec:preprocessing}

The preprocessing stage converts the source material into representations used by subsequent agents. A \emph{Docling Converter}~\cite{livathinos2025docling} first transforms each PDF into structured Markdown, preserving section hierarchy and tables. A \emph{Brief Writer} then uses $Q$ and paper digests (titles, abstracts, and section-level information) to produce a presentation brief specifying the purpose, audience, goal, tone, and level of technical detail. The brief guides subsequent planning without predetermining paper order or emphasis.

A \emph{Paper Analyst} independently processes each paper to produce an eight-field profile covering its problem, prior work, contributions, method, evidence, visuals, limitations, and a summary of its principal contributions and supporting results (Fig.~\ref{fig:contextual}). This analysis is conducted without the user request or presentation brief, ensuring that the papers are characterized before their relevance to the overall presentation is determined. Bibliographic metadata from each paper's first page, including its title, authors, affiliations, venue, and publication year is also extracted. Author names and publication years, when available, are used to construct a consistent citation registry that enables source attribution during slide content generation.

Finally, an \emph{Asset Curator} extracts relevant figures and tables, records their captions and descriptions, and organizes them into a shared asset registry containing identifiers, source provenance, captions, and explanations, allowing later agents to select visuals from any input paper.

\subsection{Outline Generation}
\label{sec:outline_generation}

\paragraph{Narrative and Content Personalization.}
When a narrative reference is provided, a \emph{Reference Analyst} examines the reference papers and their associated slide deck to infer how the presenter organizes and emphasizes multiple sources (Fig.~\ref{fig:personalization_profiles}). It identifies slide roles, paper associations, and rhetorical structure to construct two profiles: a \emph{narrative profile}, describing organization, navigation, transitions, and opening and closing patterns; and a \emph{content profile}, describing slide allocation, per-paper structure, and the use of synthesis and results. These profiles guide the subsequent analysis and outline creation. Without a reference, the framework uses default preferences derived from the presentation brief.

\begin{figure}[t]
    \centering
    \includegraphics[width=0.95\columnwidth]{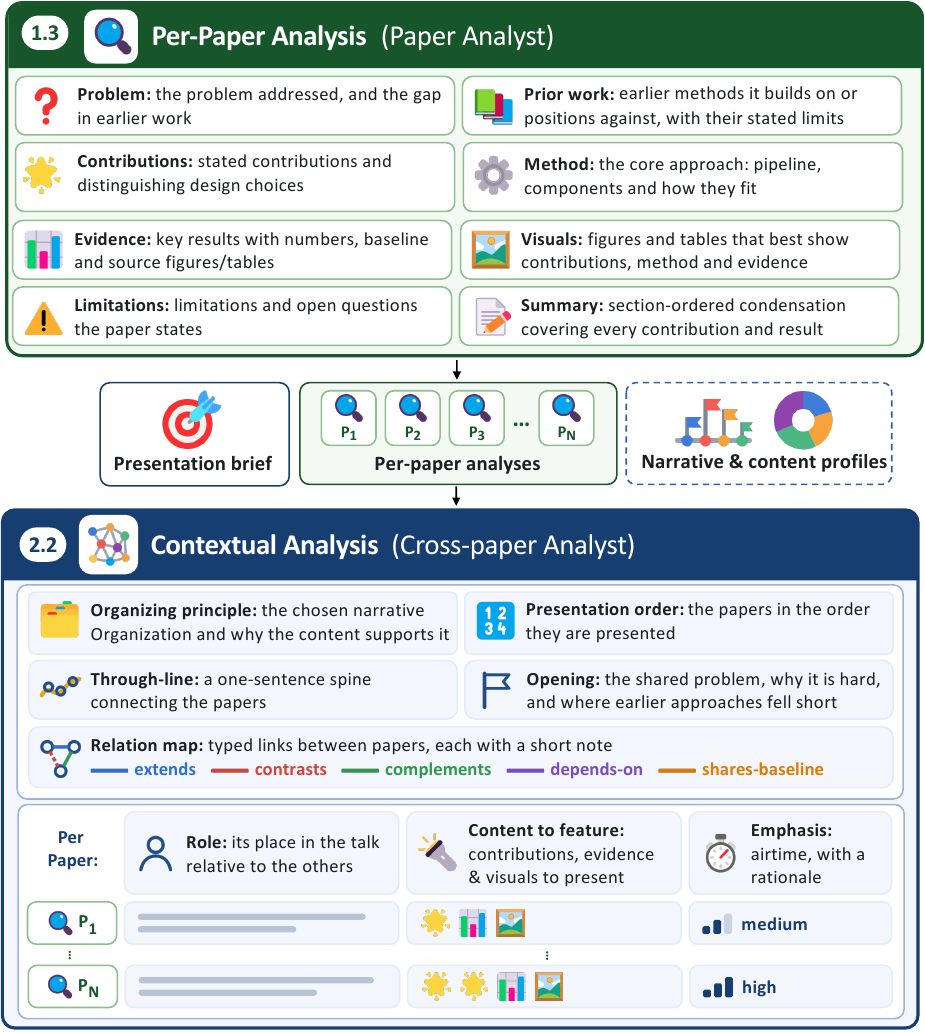}
    \vspace{-2mm}
    \caption{\textbf{Two-level contextual analysis.}
    (1) The Paper Analyst independently analyzes each source paper into eight fields covering its problem, prior work, contributions, method, evidence, visuals, limitations, and summary.
    (2) The Cross-paper Analyst integrates these analyses with the presentation brief and optional narrative and content profiles to determine the narrative structure, cross-paper relationships, paper roles, content selection, and relative emphasis.}
    \vspace{-2mm}
    \label{fig:contextual}
\end{figure}

\paragraph{Contextual Analysis.}
A \emph{Cross-paper Analyst} jointly receives the presentation brief, all per-paper analyses and summaries, conditioned on the narrative and content profiles (Fig.~\ref{fig:contextual}). At the collection level, it determines an organizing principle and presentation order, a central \emph{through-line} connecting the papers, an opening that establishes their shared problem and motivation, and a \emph{relation map} of their connections. The map uses relationships of \emph{extends}, \emph{contrasts}, \emph{complements}, \emph{depends-on}, and \emph{shares-baseline}.

At the paper level, the analyst assigns each paper a role in the overall narrative, selects the contributions, evidence, and visuals to feature, and determines its relative emphasis. When a content profile is available, its allocation preferences guide these emphasis decisions. The output is a structured \emph{contextual analysis} in JSON format that connects the papers into a coherent narrative rather than simply concatenating independent summaries. For a single input paper, its independent analysis is used directly. See Supp.~\ref{supp:example_analyses} for examples of generated per-paper and contextual analyses.

\paragraph{Outline Creation.}
A \emph{Commitment Builder} receives the contextual analysis, presentation brief, and available personalization profiles. It translates their narrative and content requirements into slide selection and high-level commitments, including the deck's section structure, distribution of slides across papers, synthesis opportunities, and narrative progression. When personalization is enabled, these commitments reflect the reference's navigation patterns, per-paper structure, and allocation preferences. An \emph{Outline Writer} then converts the commitments into a slide-by-slide outline, specifying its title, key content, and intended role. The commitments and outline jointly provide the narrative and content specification for slide generation.

\paragraph{Conversational Outline Refinement.}
The user reviews the outline and can provide free-text feedback on its content, ordering, or structure. The \emph{Outline Reviser} follows the ReSpAct mechanism~\cite{dongre2024respactharmonizingreasoningspeaking}, interleaving \emph{reasoning} about the requested changes, \emph{speaking} to clarify ambiguities when necessary, and \emph{acting} through structured editing or retrieval tools. The agent receives the current outline, presentation brief, feedback, and relevant paper material. It can modify, add, remove, reorder, split, or merge slides without regenerating unaffected portions of the outline. Each tool action returns an observation that informs the next step, enabling the agent to correct failed edits. When requested by the user, it may also retrieve additional background from the web or arXiv. The revised outline is passed to slide generation.

\subsection{Slide Generation}
\label{sec:slide_generation}

\begin{figure}[t]
    \centering
    \includegraphics[width=\columnwidth]{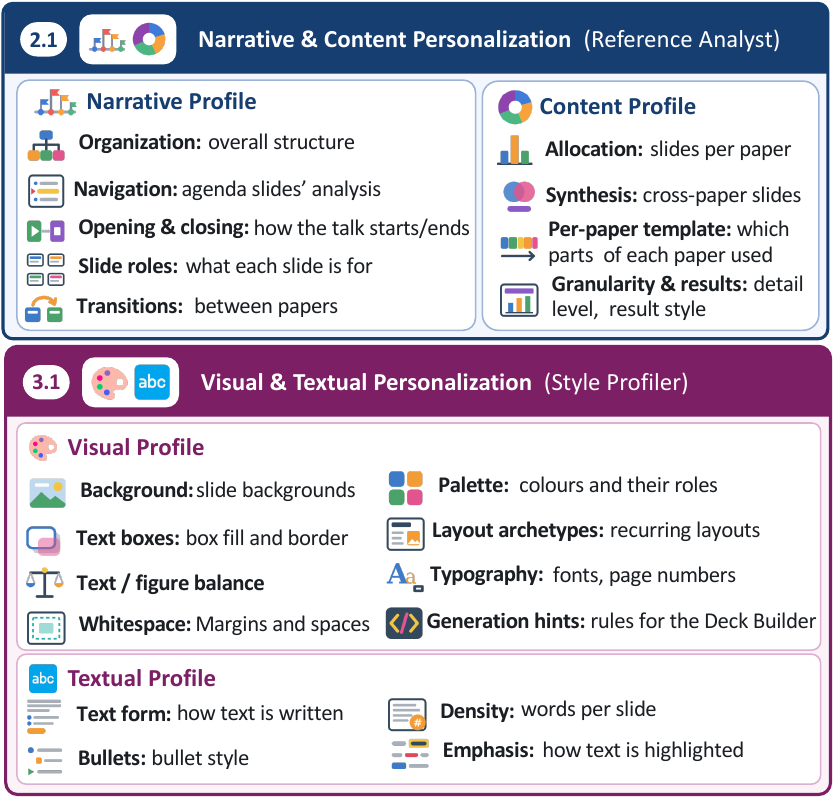}
    \vspace{-7mm}
    \caption{\textbf{Overview of the personalization profiles.} The profiles capture personalization details across four dimensions: narrative, content, visual, and textual. Profile attributes are represented as categorical choices, quantitative measurements, or natural-language descriptions. See Supp.~\ref{supp:example_profiles} for example profiles.}
    \label{fig:personalization_profiles}
\end{figure}

\paragraph{Visual and Textual Personalization.}
A \emph{Style Profiler} analyzes the optional visual reference deck to extract visual and textual preferences (Fig.~\ref{fig:personalization_profiles}). Font families and sizes, color palettes, element positions, margins, whitespace, figure-to-text ratios, and text density are extracted from text and vector layers and rendered slides. A VLM captures higher-level conventions, including recurring layouts, color roles, and emphasis patterns.

The resulting \emph{visual profile} describes backgrounds, typography, layouts, graphical treatments, and the balance between text and figures. The \emph{textual profile} describes preferred forms of expression (\textit{e.g.,} paragraphs, bullets, or short phrases), density, and emphasis conventions. These profiles inform the slide content budget, visual selection, deck construction, and subsequent refinement. Without a visual reference, the system applies a default design style. 

\paragraph{Slide Generation and Refinement.}
A \emph{Figure Matcher} receives the outline and asset registry and identifies relevant figures and tables by comparing slide content with asset captions and explanations. It then assigns supporting assets across the deck, allowing synthesis slides to draw visual evidence from multiple papers. A \emph{Content Planner} combines the outline, source texts, figure matches, and personalization profiles to produce a \emph{slide content plan} specifying slide text, source-grounded citations, and visual placements.

A \emph{Deck Builder} transforms the content plan into an HTML presentation, using the visual and textual profiles to guide layouts, text density, typography, and overall consistency. A \emph{Deck Refiner} then uses rendered-slide measurements and reference-guided visual inspection to address issues such as text overflow, poor contrast, excessive whitespace, and inappropriate figure sizing. Refinements are constrained to preserve the intended slide content and visual assets. The output is a refined deck ready for user review and PowerPoint export.

\paragraph{Conversational Slide Feedback.}
The user can inspect the rendered slides and request changes to their content, organization, or appearance. Following the same ReSpAct mechanism, the \emph{Slide Reviser} reasons about the feedback, asks clarification questions when necessary, and acts through targeted slide-level edits or deck-wide styling operations. It receives the HTML deck, the outline as a source of content, and the user's feedback. Observations from each editing operation guide subsequent actions, while unmodified slides are preserved. The resulting deck is exported as an editable PowerPoint presentation with dom-to-pptx~\cite{jagtap_domtopptx}.

\begin{figure}[t]
    \centering
    \includegraphics[width=\columnwidth]{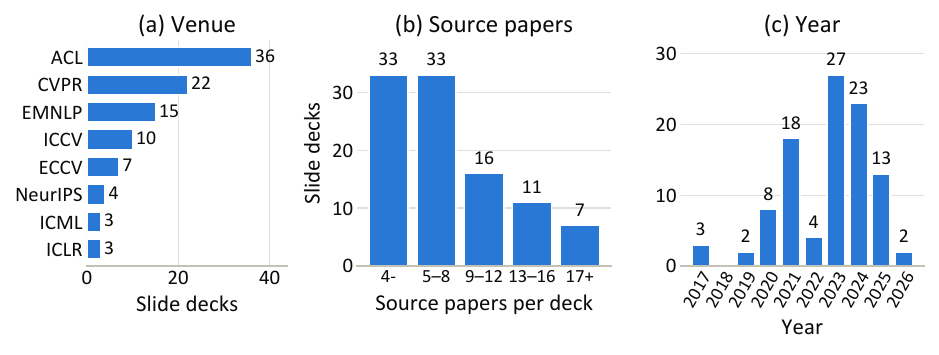}
    \vspace{-7mm}
    \caption{\textbf{Overview of KeyBench.}
    Distribution of slide decks across (a) conference venues, (b) number of source papers per deck, and (c) presentation years.}
    \label{fig:KeyBench}
    \vspace{-3mm}
\end{figure}

\section{KeyBench}
KeyBench comprises 100 source-paper sets paired with presentations collected from tutorials and keynotes at major computer vision, machine learning, and natural language processing conferences and their associated workshops, from the venues of CVPR, ICCV, ECCV, ICLR, NeurIPS, ICML, ACL, and EMNLP. Fig.~\ref{fig:KeyBench} illustrates the distribution of KeyBench across venues, the number of source papers per presentation, and presentation years. We also release the full collection of 397 slide decks gathered from the same venues, including 297 for which complete sets of source papers could not be retrieved, to facilitate further research on scientific presentation generation and analysis. Additional statistics are provided in Supp.~\ref{supp:keybench_details}.

\input{tables/keybench_eval}

\section{Experiments}

\subsection{Experimental Setup}

\paragraph{Evaluation Protocols.}
We evaluate \methodname{} in four settings:
(i) multi-paper-to-slide generation (multi-PTSG) quality,
(ii) single-paper-to-slide generation (single-PTSG) quality,
(iii) personalization quality, and
(iv) conversational goal satisfaction.
For multi-PTSG, we evaluate 100 samples from KeyBench, while for single-PTSG, we adopt the ArcBench benchmark~\cite{ozden2026narrative}, consisting of 100 research papers.

For personalization, we randomly select 20 keynote and 20 tutorial presentations as reference decks and use the source papers of the remaining 30 keynotes and 30 tutorials for generation. Each sample is paired with randomly selected narrative and visual references from the same presentation category. For conversational evaluation, we adapt the protocol from~\cite{ozden2026convdeck} to the multi-paper setting, using source-paper sets of 20 corresponding KeyBench presentations.

\paragraph{Baselines.}
We compare \methodname{} against different baselines for each evaluation setting:
(i) for multi-PTSG quality, SlideGen~\cite{liang2025slidegen}, ArcDeck~\cite{ozden2026narrative}, EvoPresent~\cite{liu2025presenting}, ConvDeck~\cite{ozden2026convdeck}, and DeepPresenter~\cite{zheng2026deeppresenter};
(ii) for single-PTSG quality, the same methods together with SlideTailor~\cite{zeng2026slidetailor};
(iii) for personalization quality, SlideTailor; and
(iv) for conversational goal satisfaction, ConvDeck and ConvHTML, a conversational HTML-based slide generation baseline with post-generation refinement.

\paragraph{Evaluation Metrics.}
For slide generation quality, we employ two complementary evaluation approaches: rubric-based assessment and question answering. The rubric-based evaluation measures four dimensions: (i) narrative flow, (ii) textual quality, (iii) visual thematic consistency, and (iv) visual layout. Each dimension is assessed using a structured grading rubric with scores ranging from 0 to 10.

To assess the content conveyed by generated decks, we introduce a multi-paper question-answering evaluation. For each sample, GPT-5 generates 12 questions across four categories, with three questions each: (i) \emph{Story}, probing the talk's motivation, central thesis, and cross-paper connections; (ii) \emph{Visuals}, requiring information from figures, plots, tables, or diagrams; (iii) \emph{Hard}, targeting reasoning beyond surface content, such as trade-offs and cross-method comparisons; and (iv) \emph{Depth}, assessing technical understanding of methods and results. Questions are generated from the source papers and the human-authored reference deck, which guides the selection of presentation-worthy content rather than serving as an answer source. Each question is paired with a reference answer and key points. The evaluator answers each question using only the generated deck, and answers are graded as correct (2), partial (1), or missing/incorrect (0). Per-category scores are averaged and scaled to 0--100. The same questions are used for all methods. For single-PTSG, we follow the evaluation protocol of ArcBench~\cite{ozden2026narrative}.

We additionally conduct pairwise comparisons to assess overall generation quality and personalization quality. For personalization, we evaluate adherence to reference presentations across four dimensions: narrative, content, visual, and textual, together with an overall assessment. For conversational quality, we measure user goal
satisfaction across five goal inventories:
(i) Narrative Structure,
(ii) Content Inclusion/Exclusion,
(iii) Deck Composition,
(iv) Figure and Table Usage, and
(v) Style and Wording.
We adapt the goals in each inventory to reflect
multi-PTSG requirements and compare methods
based on their goal satisfaction.  Finally, computational efficiency is assessed through token consumption, inference cost, and runtime. See Supp.~\ref{supp:eval_details} for evaluation rubrics, prompts, and goal inventories.

\paragraph{Implementation Details.}
For fair comparisons, all LLM-based methods use GPT-5~\cite{singh2025openai_gpt5}  minimal reasoning and GLM5.3-Flash~\cite{glm5team2026glm5vibecodingagentic} with low effort. We disable both personalization and conversational refinement when evaluating overall generation quality. To apply single-PTSG baselines to the multi-PTSG setting, we concatenate the source papers after removing their bibliographies and appendices. For personalization experiments, we disable conversational refinement, while for conversational evaluation, we disable personalization. To enable visual personalization in SlideTailor, we adapt the method to generate a PPTX template from the visual reference deck and use it as input for slide generation. The ConvHTML prompt and further implementation details are provided in Supp.~\ref{supp:implementation_details}.

\subsection{Results and Analysis}

\begin{figure}[t]
    \centering
    \includegraphics[width=\columnwidth]{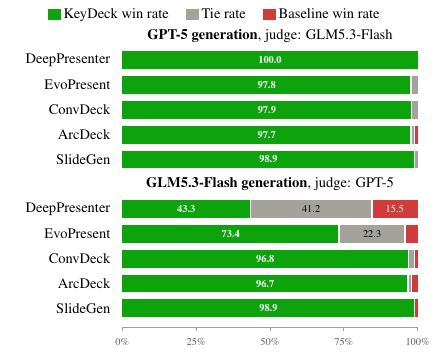}
    \vspace{-9mm}
    \caption{\textbf{Pairwise comparison.} Proportion of generated decks where \methodname{} wins (green), the two presentation orders disagree (tie, gray), or the baseline wins (red), on overall presentation quality. Each pair is judged twice with the order swapped.}
    \label{fig:pairwise}
    \vspace{-2mm}
\end{figure}

\paragraph{(i) Slide Generation Quality.}
As shown in Tab.~\ref{tab:keybench_eval}, \methodname{} achieves the highest average scores in both reference-based Q/A and rubric-based quality assessment across both generation backbones and evaluators. Pairwise comparisons (Fig.~\ref{fig:pairwise}) further show win rates above 97\% against all baselines with GPT and win-or-tie rates above 84\% with GLM backbone. In the single-paper setting (Tab.~\ref{tab:arcbench_eval}), \methodname{} also achieves the highest average rubric-based quality, demonstrating its effectiveness across both settings. Additional qualitative examples are provided in Supp.~\ref{supp:qual_examples}.

\input{tables/arcbench_eval}

\paragraph{(ii) Personalization Quality.}
Fig.~\ref{fig:pairwise_pers_arcdeck_slidetailor} shows that \methodname{} is consistently preferred over SlideTailor across all four personalization dimensions and in overall personalization quality under both generation settings. In overall quality, \methodname{} achieves win rates of 55.0\% and 65.0\% with GPT and GLM generation, compared to 11.7\% for SlideTailor in both settings. These results demonstrate the effectiveness of our reference-based personalization approach.

\input{tables/personalization_eval}

\paragraph{(iii) Conversational Quality.}
As shown in Tab.~\ref{tab:user_goal_satisfaction}, \methodname{} outperforms all conversational baselines overall, while also achieving a high user goal satisfaction rate of 84\%. These results indicate the effectiveness of its conversational refinement capabilities in adapting multi-paper presentations to user feedback.

\input{tables/conversation_eval}

\paragraph{(iv) Ablation Study.}
We investigate the contributions of contextual analysis and per-paper analysis through the ablation study presented in Tab.~\ref{tab:ablation}. Removing either component leads to decreased performance in story- and depth-related question answering, as well as narrative flow. These findings highlight the importance of both paper-level understanding and cross-paper reasoning for producing coherent and informative multi-paper presentations.

\input{tables/ablation}

\begin{figure}[t]
  \centering
  \includegraphics[width=0.9\columnwidth]{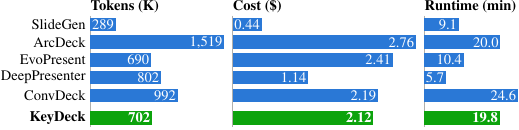}
  \vspace{-3mm}
  \caption{\textbf{Average generation cost and runtime.} Averaged over a set of 10 talks, with the generation backbone of GPT-5.}
  \label{fig:gen_cost_runtime}
  \vspace{-4mm}
\end{figure}

\paragraph{(v) Computational Efficiency.}
As shown in Fig.~\ref{fig:gen_cost_runtime}, for multi-PTSG, \methodname{} ranks third in runtime and fourth in token consumption and inference cost among six methods (highest to lowest), while achieving superior generation quality at comparable computational cost.

\section{Conclusion}

We introduce \methodname, a framework for multi-paper-to-slide generation that integrates structured cross-paper synthesis with reference-guided personalization. By extracting narrative, content, visual, and textual preferences from reference presentations, \methodname adapts both the organization and presentation of information to users' preferences. We also introduce \textit{KeyBench}, a benchmark of 100 expert-authored multi-paper presentations, along with evaluation protocols for generation quality, personalization adherence, and conversational goal satisfaction. Extensive experiments demonstrate that \methodname outperforms existing approaches in both multi-paper and single-paper slide generation, achieves stronger personalization adherence while maintaining comparable computational costs.

\clearpage
\newpage

{
    \small
    \bibliographystyle{ieeenat_fullname}
    \bibliography{main}
}

\input{sec/X_suppl}

\end{document}

%% file: preamble.tex
\usepackage{cuted} 

\usepackage{currfile} 

\usepackage{caption} 

\usepackage{booktabs}
\usepackage{multirow}
\usepackage{graphicx}
\usepackage[table]{xcolor}
\usepackage{tabularx}
\usepackage{titletoc}

\usepackage{tabularx}
\usepackage{colortbl}
\usepackage{pifont}
\usepackage{float}


%% file: tables/keybench_eval.tex
\begin{table*}[!t]

\caption{\textbf{KeyBench evaluation.}
Comparison of \textbf{multi-PTSG} methods using reference-based Q/A and rubric-based quality assessment.
Average denotes the mean of four metrics per assessment group.
\textbf{Bold} and \underline{underlined} values indicate the best
and second-best results.}
\label{tab:keybench_eval}
  \vspace{-2mm}
  \centering
  \small
  \setlength{\tabcolsep}{4pt}
  \renewcommand{\arraystretch}{1.15}
  \resizebox{\textwidth}{!}{%
  \begin{tabular}{@{}l*{20}{c}@{}}
  \toprule

  \multirow{3}{*}{\textbf{Method}}
  & \multicolumn{10}{c}{\textbf{Multi-PTSG Reference-based Q/A (0--100)}}
  & \multicolumn{10}{c}{\textbf{Multi-PTSG Rubric-based Assessment (0--10)}} \\

  \cmidrule(lr){2-11}
  \cmidrule(lr){12-21}

  & \multicolumn{2}{c}{\textbf{Story}}
  & \multicolumn{2}{c}{\textbf{Visuals}}
  & \multicolumn{2}{c}{\textbf{Hard}}
  & \multicolumn{2}{c}{\textbf{Depth}}
  & \multicolumn{2}{c}{\textbf{Average}}
  & \multicolumn{2}{c}{\textbf{Narrative Flow}}
  & \multicolumn{2}{c}{\textbf{Textual Quality}}
  & \multicolumn{2}{c}{\textbf{Visual Thematic}}
  & \multicolumn{2}{c}{\textbf{Visual Layout}}
  & \multicolumn{2}{c}{\textbf{Average}} \\

  \cmidrule(lr){2-3}
  \cmidrule(lr){4-5}
  \cmidrule(lr){6-7}
  \cmidrule(lr){8-9}
  \cmidrule(lr){10-11}
  \cmidrule(lr){12-13}
  \cmidrule(lr){14-15}
  \cmidrule(lr){16-17}
  \cmidrule(lr){18-19}
  \cmidrule(lr){20-21}

  & GPT & GLM
  & GPT & GLM
  & GPT & GLM
  & GPT & GLM
  & GPT & GLM
  & GPT & GLM
  & GPT & GLM
  & GPT & GLM
  & GPT & GLM
  & GPT & GLM \\

  \midrule

  \multicolumn{21}{c}{\cellcolor{RoyalBlue!8}\textbf{Generation Model: GPT-5}} \\

  SlideGen~\cite{liang2025slidegen}
  & 25.1 & 27.4 & 23.6 & 23.6 & 27.4 & 27.2 & 31.6 & 34.6
  & 26.93 & 28.20
  & 6.30 & 2.75 & 5.56 & \underline{6.39} & 6.86 & 5.58 & 8.51 & \textbf{6.94}
  & 6.81 & \underline{5.42} \\

  ArcDeck~\cite{ozden2026narrative}
  & \underline{57.2} & 52.5 & \underline{34.2} & \underline{41.6} & 39.2 & 44.7 & 45.6 & 50.4
  & 44.05 & \underline{47.30}
  & 6.94 & 2.86 & 5.44 & 5.47 & \underline{8.05} & \underline{6.56} & 8.28 & 6.09
  & 7.18 & 5.25 \\

  EvoPresent~\cite{liu2025presenting}
  & \underline{57.2} & \underline{57.6} & 18.6 & 18.8 & \underline{50.6} & \underline{50.6} & \underline{56.5} & \underline{55.9}
  & \underline{45.73} & 45.73
  & \underline{8.47} & \underline{4.37} & \underline{7.11} & 5.77 & 6.68 & 2.85 & \textbf{9.84} & 5.00
  & \underline{8.03} & 4.50 \\

  ConvDeck~\cite{ozden2026convdeck}
  & 54.9 & 53.6 & 28.7 & 34.8 & 46.6 & 46.8 & 49.8 & 53.2
  & 45.00 & 47.10
  & 6.65 & 1.90 & 5.78 & 4.90 & 7.72 & 3.95 & 8.34 & 5.68
  & 7.12 & 4.11 \\

  DeepPresenter~\cite{zheng2026deeppresenter}
  & 37.8 & 41.4 & 5.7 & 7.6 & 8.2 & 12.9 & 11.0 & 16.7
  & 15.67 & 19.65
  & 6.05 & 3.58 & 5.23 & 5.16 & 2.24 & 1.77 & 8.77 & \underline{6.68}
  & 5.57 & 4.30 \\

  \rowcolor{gray!15}
  \methodname
  & \textbf{81.6} & \textbf{80.2} & \textbf{51.7} & \textbf{53.6} & \textbf{76.4} & \textbf{78.9} & \textbf{78.1} & \textbf{82.9}
  & \textbf{71.95} & \textbf{73.90}
  & \textbf{9.89} & \textbf{9.01} & \textbf{7.87} & \textbf{9.04} & \textbf{9.62} & \textbf{7.05} & \underline{9.44} & 6.47
  & \textbf{9.21} & \textbf{7.89} \\

  \addlinespace[2pt]
  \multicolumn{21}{c}{\cellcolor{ForestGreen!8}\textbf{Generation Model: GLM5.3-Flash}} \\

  SlideGen~\cite{liang2025slidegen}
  & 18.3 & 22.6 & 17.1 & 21.7 & 20.7 & 24.0 & 24.4 & 29.1
  & 20.12 & 24.35
  & 6.01 & 3.27 & 4.80 & 5.50 & 6.41 & 4.99 & 8.17 & 5.55
  & 6.35 & 4.83 \\

  ArcDeck~\cite{ozden2026narrative}
  & 62.8 & 64.8 & \underline{41.5} & \underline{48.8} & 57.3 & 62.8 & 61.8 & 66.3
  & 55.85 & 60.67
  & 7.15 & 4.35 & 7.09 & 7.80 & 8.94 & \underline{7.37} & 7.22 & 4.80
  & 7.60 & 6.08 \\

  EvoPresent~\cite{liu2025presenting}
  & 68.7 & 71.7 & 38.6 & 38.0 & \underline{70.5} & \underline{72.6} & \underline{73.0} & \underline{75.8}
  & \underline{62.70} & 64.53
  & 9.38 & 8.28 & \underline{7.93} & \underline{8.49} & 8.99 & 2.29 & 9.21 & 5.48
  & 8.88 & 6.13 \\

  ConvDeck~\cite{ozden2026convdeck}
  & 68.9 & 70.9 & 38.0 & 46.7 & 65.2 & 71.3 & 67.9 & 73.2
  & 60.00 & \underline{65.53}
  & 7.46 & 4.68 & 6.85 & 8.17 & 8.46 & 6.54 & 8.44 & 6.26
  & 7.80 & 6.41 \\

  DeepPresenter~\cite{zheng2026deeppresenter}
  & \underline{71.7} & \underline{77.2} & 29.9 & 30.7 & 50.0 & 62.4 & 57.5 & 66.9
  & 52.27 & 59.30
  & \underline{9.66} & \underline{9.21} & 7.16 & 7.87 & \underline{9.00} & 6.98 & \textbf{9.90} & \textbf{8.43}
  & \underline{8.93} & \underline{8.12} \\

  \rowcolor{gray!15}
  \methodname
  & \textbf{77.0} & \textbf{77.8} & \textbf{48.6} & \textbf{52.4} & \textbf{73.6} & \textbf{76.0} & \textbf{78.7} & \textbf{81.5}
  & \textbf{69.47} & \textbf{71.92}
  & \textbf{9.89} & \textbf{9.85} & \textbf{8.01} & \textbf{9.27} & \textbf{9.80} & \textbf{8.51} & \underline{9.66} & \underline{7.73}
  & \textbf{9.34} & \textbf{8.84} \\

  \bottomrule
  \end{tabular}%
  }
  \vspace{-2mm}
  \end{table*}

%% file: tables/arcbench_eval.tex
\begin{table*}[t!]
\centering
\small
\setlength{\tabcolsep}{4pt}
\renewcommand{\arraystretch}{1.15}
\caption{\textbf{ArcBench evaluation.}
Comparison of \textbf{single-PTSG} methods using reference-based Q/A
and rubric-based quality assessment.
Average denotes the mean of four metrics per assessment group.
\textbf{Bold} and \underline{underlined} values indicate the best
and second-best results.}
\label{tab:arcbench_eval}
\vspace{-2mm}
\resizebox{\textwidth}{!}{%
\begin{tabular}{@{}l*{20}{c}@{}}
\toprule

\multirow{3}{*}{\textbf{Method}}
& \multicolumn{10}{c}{\textbf{Single-PTSG Reference-based Q/A}}
& \multicolumn{10}{c}{\textbf{Single-PTSG Rubric-based Assessment}} \\

\cmidrule(lr){2-11}
\cmidrule(lr){12-21}

& \multicolumn{2}{c}{\textbf{Story}}
& \multicolumn{2}{c}{\textbf{Visuals}}
& \multicolumn{2}{c}{\textbf{Hard}}
& \multicolumn{2}{c}{\textbf{Depth}}
& \multicolumn{2}{c}{\textbf{Average}}
& \multicolumn{2}{c}{\textbf{Narrative Flow}}
& \multicolumn{2}{c}{\textbf{Textual Quality}}
& \multicolumn{2}{c}{\textbf{Visual Thematic}}
& \multicolumn{2}{c}{\textbf{Visual Layout}}
& \multicolumn{2}{c}{\textbf{Average}} \\

\cmidrule(lr){2-3}
\cmidrule(lr){4-5}
\cmidrule(lr){6-7}
\cmidrule(lr){8-9}
\cmidrule(lr){10-11}
\cmidrule(lr){12-13}
\cmidrule(lr){14-15}
\cmidrule(lr){16-17}
\cmidrule(lr){18-19}
\cmidrule(lr){20-21}

& GPT & GLM
& GPT & GLM
& GPT & GLM
& GPT & GLM
& GPT & GLM
& GPT & GLM
& GPT & GLM
& GPT & GLM
& GPT & GLM
& GPT & GLM \\

\midrule

\multicolumn{21}{c}{\cellcolor{RoyalBlue!8}\textbf{Generation Model: GPT-5}} \\

SlideGen
& \textbf{96.2} & \textbf{99.0}
& 51.4 & 58.0
& \underline{73.5} & \underline{87.9}
& \underline{72.8} & \underline{87.2}
& 73.5 & 83.0
& 5.66 & 5.08
& 3.97 & 3.60
& 7.67 & 7.57
& 8.86 & 7.88
& 6.54 & 6.03 \\

SlideTailor
& 70.3 & 96.1
& 48.1 & 58.6
& 24.2 & 63.5
& 24.4 & 56.7
& 41.8 & 68.7
& 5.12 & 4.40
& 2.31 & 1.96
& 8.01 & \textbf{8.54}
& 8.73 & 7.09
& 6.04 & 5.50 \\

ArcDeck
& 93.4 & 98.1
& \textbf{69.3} & \textbf{70.1}
& 70.9 & 85.4
& 69.5 & 83.8
& \underline{75.8} & \underline{84.4}
& \underline{6.38} & \underline{6.14}
& 4.58 & 4.30
& \textbf{8.31} & 8.30
& 8.51 & 7.22
& 6.94 & 6.49 \\

EvoPresent
& 89.6 & 98.0
& 53.9 & 62.6
& 65.0 & 81.7
& 60.6 & 79.6
& 67.3 & 80.5
& 4.81 & 5.22
& 4.67 & 4.64
& 7.51 & 7.93
& 9.29 & 7.84
& 6.57 & 6.41 \\

ConvDeck
& 92.0 & 98.3
& \underline{64.3} & \underline{68.4}
& 62.3 & 82.8
& 61.5 & 82.3
& 70.0 & 83.0
& 4.88 & 4.68
& 3.62 & 3.64
& 8.14 & 8.26
& \underline{9.30} & \underline{8.24}
& 6.49 & 6.21 \\

DeepPresenter
& 87.8 & 97.0
& 61.6 & 68.2
& 57.4 & 79.2
& 58.5 & 79.4
& 66.3 & 81.0
& 5.84 & 6.04
& \underline{5.48} & \underline{5.48}
& 7.62 & 7.84
& 9.08 & 7.66
& \underline{7.00} & \underline{6.75} \\

\rowcolor{gray!15}
\methodname
& \textbf{96.2} & \textbf{99.0}
& 62.6 & 68.1
& \textbf{86.6} & \textbf{93.0}
& \textbf{85.2} & \textbf{91.9}
& \textbf{82.7} & \textbf{88.0}
& \textbf{7.26} & \textbf{7.27}
& \textbf{6.79} & \textbf{6.99}
& \underline{8.30} & \underline{8.36}
& \textbf{9.49} & \textbf{8.58}
& \textbf{7.96} & \textbf{7.80} \\

\addlinespace[2pt]
\multicolumn{21}{c}{\cellcolor{ForestGreen!8}\textbf{Generation Model: GLM5.3-Flash}} \\

SlideGen
& 96.9 & \underline{99.3}
& 52.0 & 58.3
& 79.9 & \underline{90.0}
& 80.7 & 89.0
& 77.4 & 84.2
& 6.89 & 6.86
& 5.91 & 6.00
& 7.73 & 7.60
& 8.74 & 7.09
& 7.32 & 6.89 \\

SlideTailor
& 71.8 & 93.5
& 51.0 & 61.2
& 26.6 & 68.8
& 29.1 & 63.7
& 44.6 & 71.8
& 5.46 & 5.09
& 2.88 & 2.61
& 7.77 & 7.98
& 8.17 & 6.94
& 6.07 & 5.66 \\

ArcDeck
& 94.8 & \underline{99.3}
& 65.5 & 69.7
& \underline{80.8} & 88.4
& \underline{81.8} & 88.3
& \underline{80.7} & \underline{86.4}
& 6.02 & 6.53
& 5.05 & 5.53
& 7.79 & 7.90
& 9.41 & 8.45
& 7.07 & 7.10 \\

EvoPresent
& 94.0 & 98.5
& 63.4 & 70.1
& 74.1 & 85.9
& 74.2 & 84.1
& 76.4 & 84.7
& \underline{7.42} & \underline{8.43}
& \underline{7.18} & \textbf{7.68}
& \underline{8.49} & \underline{8.80}
& 9.00 & 7.95
& 8.02 & 8.21 \\

ConvDeck
& \underline{97.1} & 99.1
& 60.9 & 67.2
& 79.1 & 87.5
& 80.4 & \underline{89.1}
& 79.4 & 85.7
& 6.77 & 7.33
& 5.67 & 6.29
& 7.73 & 7.85
& 9.34 & 9.00
& 7.38 & 7.62 \\

DeepPresenter
& 92.2 & 97.8
& \underline{66.6} & \underline{71.2}
& 66.5 & 82.8
& 66.1 & 82.1
& 72.9 & 83.5
& 7.01 & 7.53
& 6.96 & 7.42
& 8.44 & 8.74
& \underline{9.74} & \underline{9.22}
& \underline{8.04} & \underline{8.23} \\

\rowcolor{gray!15}
\methodname
& \textbf{98.1} & \textbf{99.6}
& \textbf{74.2} & \textbf{75.2}
& \textbf{90.9} & \textbf{94.0}
& \textbf{91.1} & \textbf{94.6}
& \textbf{88.6} & \textbf{90.9}
& \textbf{8.12} & \textbf{8.47}
& \textbf{7.38} & \underline{7.61}
& \textbf{9.07} & \textbf{9.20}
& \textbf{9.84} & \textbf{9.52}
& \textbf{8.60} & \textbf{8.70} \\

\bottomrule
\end{tabular}%
}
\vspace{-3mm}
\end{table*}

%% file: tables/personalization_eval.tex
\begin{figure}[t]
    \centering
    \includegraphics[width=0.85\columnwidth]{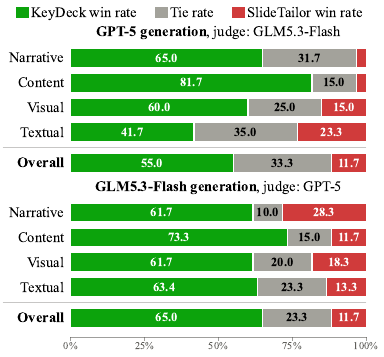}
    \vspace{-3mm}
    \caption{\textbf{Pairwise personalization judgments.}
    Proportion of generated decks where \methodname{} is preferred over SlideTailor (green), the result is a tie (gray), or SlideTailor is preferred (red). Each pair is evaluated in both orders; a win requires agreement.}
    \label{fig:pairwise_pers_arcdeck_slidetailor}
    \vspace{-2mm}
\end{figure}

%% file: tables/conversation_eval.tex
\begin{table}[h]
\centering
\scriptsize
\setlength{\tabcolsep}{3pt}
\renewcommand{\arraystretch}{1.10}
\caption{\textbf{User goal satisfaction} (\%).
We evaluate 20 samples for multi-PTSG across five categories:
Narrative Structure (NS), Content Inclusion/Exclusion (CI/E),
Deck Composition (DC), Figure/Table Usage (FTU),
and Style and Wording (SW).
All methods use GPT-5 as the backbone and GLM5.3-Flash as the judge.}
\label{tab:user_goal_satisfaction}
\vspace{-2mm}
\resizebox{0.8\columnwidth}{!}{%
\begin{tabular}{@{}l*{5}{c}|c@{}}
\toprule

\textbf{Method}
& \textbf{NS}
& \textbf{CI/E}
& \textbf{DC}
& \textbf{FTU}
& \textbf{SW}
& \textbf{Overall} \\

\midrule

ConvHTML
 & 85.0 & 80.0 & 70.0 & 45.0 & \textbf{85.0} & 73.0 \\

ConvDeck
& 75.0 & 80.0 & 70.0 & 45.0 & 80.0 & 70.0 \\

\rowcolor{gray!15}
\methodname
& \textbf{90.0} & \textbf{85.0} & \textbf{80.0} & \textbf{85.0} & 80.0 & \textbf{84.0} \\

\bottomrule
\end{tabular}%
}
\vspace{-5mm}
\end{table}

%% file: tables/ablation.tex
\begin{table}[t]
\caption{\textbf{Ablation study.} \methodname{} and its ablations on 10 GPT-5-generated samples, evaluated by GLM.}
\label{tab:ablation}
\vspace{-2mm}
\centering
\small
\setlength{\tabcolsep}{4pt}
\resizebox{0.7\columnwidth}{!}{%
\begin{tabular}{@{}lccc@{}}
\toprule
\textbf{Variant} & \textbf{Story} & \textbf{Depth} & \textbf{NF} \\
\midrule
\rowcolor{gray!15}
\methodname{} & 78.3 & 83.3 & 9.00 \\
\quad w/o contextual analysis & 73.3 & 81.7 & 8.40 \\
\quad w/o per-paper analysis & 71.7 & 75.0 & 8.40 \\
\bottomrule
\end{tabular}%
}
\end{table}

%% file: sec/X_suppl.tex
\clearpage
\appendix
\setcounter{page}{1}
\maketitlesupplementary

\section*{Appendix Contents}

\startcontents[appendices]
\printcontents[appendices]{}{1}{}


\section{User Studies}
\label{supp:user_study}

\subsection{Multi-PTSG User Study}

\begin{figure*}[t]
    \centering
    \includegraphics[width=\textwidth]{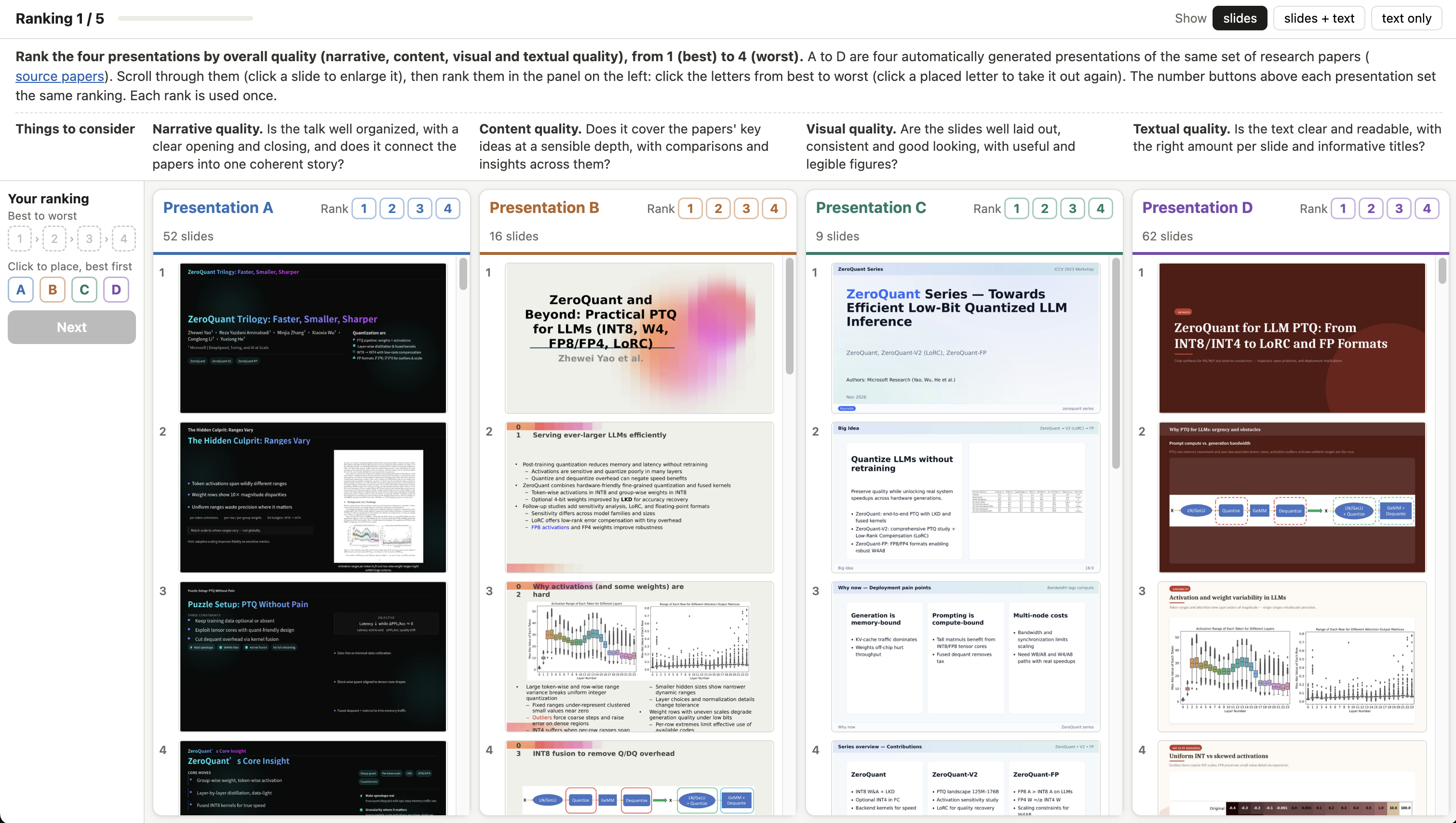}
    \caption{\textbf{User study interface for multi-PTSG quality ranking evaluation.}
    Participants review four anonymized presentations generated from the same set of research papers using the GPT-5 backbone and rank them from 1 (best) to 4 (worst) based on overall narrative, content, visual, and textual quality.}
    \label{fig:user_study_multiptsg_ui}
\end{figure*}

To complement the automated evaluations in the main paper, we conduct a human ranking study to assess the overall quality of multi-PTSG generations. Participants are asked to rank the presentations generated by KeyDeck, ArcDeck, EvoPresent, and DeepPresenter using the GPT-5 backbone and the same set of source papers from 1 (best) to 4 (worst), considering narrative, content, visual, and textual quality. We recruit 30 participants through Prolific, each completing five ranking tasks, resulting in 150 rankings equally distributed between keynotes and tutorials. All slide decks were displayed in randomized order, and in scrollable interfaces, allowing participants to inspect every slide, while the source papers could be accessed through external links.  The user study interface is shown in Fig.~\ref{fig:user_study_multiptsg_ui}.

As shown in Table~\ref{tab:human_ranking}, KeyDeck achieves the best average rank of 1.61, outperforming ArcDeck (2.33), EvoPresent (2.77), and DeepPresenter (3.29). KeyDeck also achieves the best average rank for both keynotes (1.57) and tutorials (1.64), demonstrating consistent human preference across both presentation types.

\begin{table}[t]
\centering
\small
\caption{\textbf{Human ranking study on overall multi-PTSG quality.} Each respondent ranks the four systems' presentations of the same set of papers from 1 (best) to 4 (worst).}
\label{tab:human_ranking}
\begin{tabular}{l ccc}
\toprule
 & \multicolumn{3}{c}{Average rank $\downarrow$} \\
\cmidrule(lr){2-4}
Method & Overall & Keynotes & Tutorials \\
\midrule
KeyDeck (ours) & \textbf{1.61} & \textbf{1.57} & \textbf{1.64} \\
ArcDeck & 2.33 & 2.29 & 2.37 \\
EvoPresent & 2.77 & 2.77 & 2.76 \\
DeepPresenter & 3.29 & 3.36 & 3.23 \\
\bottomrule
\end{tabular}
\end{table}

\begin{table}[t]
\centering
\small
\setlength{\tabcolsep}{7pt}
\renewcommand{\arraystretch}{1.15}
\caption{\textbf{Human evaluation of personalization adherence.}
Selection rates (\%) for \methodname{} and SlideTailor
across 150 pairwise comparisons.}
\label{tab:human_eval_personalization}
\vspace{-2mm}
\begin{tabular}{@{}lccc@{}}
\toprule
\textbf{Method} &
\textbf{Overall} & 
\textbf{Keynotes} &
\textbf{Tutorials} \\
 
\midrule
\methodname{} & 
\textbf{80.7} &
\textbf{81.3} & \textbf{80.0} \\ 
SlideTailor   & 19.3 & 18.7 & 20.0 \\
\bottomrule
\end{tabular}
\end{table}

\subsection{Personalization User Study}

\begin{figure*}[t]
    \centering
    \includegraphics[width=\textwidth]{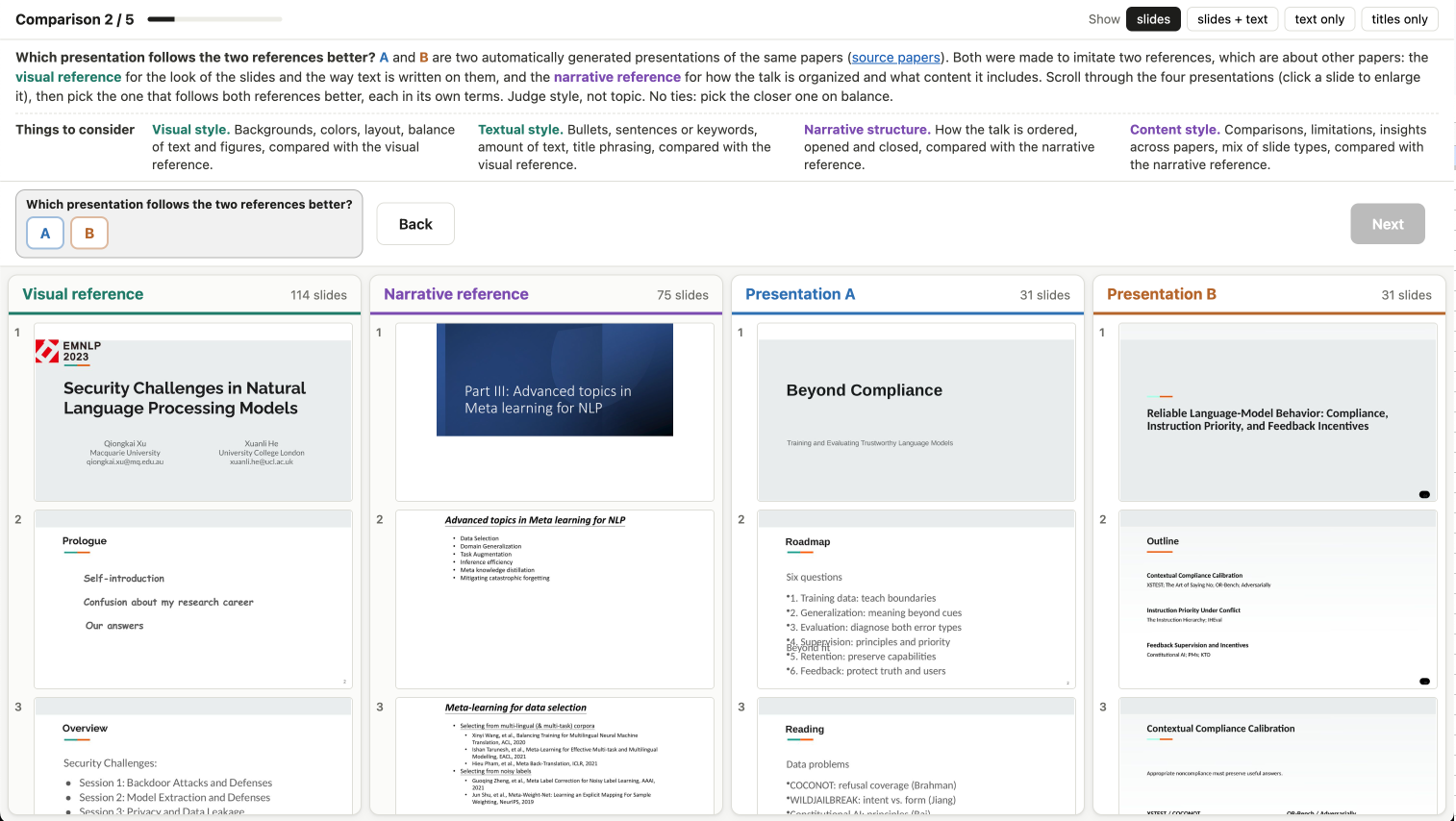}
    \caption{\textbf{User study interface for pairwise personalization evaluation.}
    Participants compare two anonymized presentations generated from the same set of research papers, narrative and visual reference slide decks to review their slides side by side, and select their preferred presentation based on overall personalization quality.}
    \label{fig:ui_pairwise_personalization}
\end{figure*}

We conduct an additional human evaluation study to assess the adherence quality to reference-based personalization preferences.
Participants compare presentations generated by \methodname{} and SlideTailor from the same source papers and personalization references, with GPT-5 backbone. The narrative reference consists of source papers and their corresponding presentation, while the visual reference provides guidance for visual and textual style. Participants evaluate adherence to the overall reference personalization. The study was completed with 30 participants recruited from Prolific. Each participant evaluated five presentation pairs, with presentation order randomized. Participants were shown the complete narrative and visual reference slide decks, along with links to the source papers and two generated presentations (A and B). All slide decks were displayed in scrollable interfaces, allowing participants to inspect every slide, while the source papers could be accessed through external links. Participants were asked to select which generated presentation better adhered to the narrative and visual references overall. The web interface of this study is shown in Fig.~\ref{fig:ui_pairwise_personalization}.

As shown in Table~\ref{tab:human_eval_personalization}, \methodname{} is preferred over SlideTailor in 80.7\% of the 150 pairwise comparisons. This preference is consistent across presentation types, with 81.3\% for keynotes and 80.0\% for tutorials. These results provide human evaluation evidence that \methodname{} better adheres to the provided personalization references than SlideTailor, regardless of presentation type.

\section{Implementation Details}
\label{supp:implementation_details}

\subsection{Baseline Adaptation and System Inputs} Since several existing baselines were originally developed for single-PTSG, we adapt their inputs to enable multi-paper generation. Specifically, for SlideGen~\cite{liang2025slidegen}, ArcDeck~\cite{ozden2026narrative}, EvoPresent~\cite{liu2025presenting}, and ConvDeck~\cite{ozden2026convdeck}, we remove the bibliographies and appendices from the source papers and concatenate their remaining content into a single input document. This allows these methods to process the complete source-paper collection without modifying their underlying generation pipelines.

In contrast, DeepPresenter~\cite{zheng2026deeppresenter} supports autonomous document retrieval and multi-source processing. We therefore provide the source-paper information directly and allow the system to retrieve and process the papers using its native workflow, without imposing document concatenation or additional restrictions.

All methods use the same source-paper sets and generation backbones for comparison.

\paragraph{System Inputs.} For all multi-PTSG methods that accept a target slide count or presentation duration, we specify 54 slides or 54 minutes, respectively, depending on the supported input format. The target slide count is based on the average number of slides per presentation in KeyBench of 53.9, while the target duration follows the common rule of thumb of approximately one minute per slide. For baselines that accept free-form user instructions, we provide the prompt \textit{Generate an academic keynote/tutorial presentation based on the input papers}, replacing keynote/tutorial with \textit{keynote} or \textit{tutorial} according to the presentation type of the corresponding reference deck. For all methods that support aspect ratio specification, we use a 16:9 aspect ratio. 

\subsection{Conversational Evaluation Details}

For conversational evaluation, we include ConvHTML, an HTML-based conversational slide generation baseline introduced in ConvDeck~\cite{ozden2026convdeck}. ConvHTML is a conversational baseline that generates and refines an HTML slide deck with three agents. The generation agent receives the user instruction and the text of all source documents and produces the complete deck in a single call. The review agent simulates the user: given the rendered slides and the goals that are not yet satisfied, it returns a verdict for each goal and a list of feedback items phrased as a user would phrase them. The reviser agent then edits the HTML according to this feedback, using the source documents as ground truth. Review and revision alternate for at most five rounds and stop early once the review agent considers every goal satisfied.

We adapt ConvHTML to multi-PTSG by providing the complete source-paper collection and modifying the generation instructions to request a unified presentation that integrates information across papers.

The complete prompt used for ConvHTML in multi-PTSG setting is provided in Fig.~\ref{fig:conv_html_prompt}.

\paragraph{Goal Inventories.}
Fig.~\ref{fig:conv_goal_inventory} lists the five goal inventories,  Narrative Structure (NS), Content Inclusion/Exclusion (CI/E),
Deck Composition (DC), Figure/Table Usage (FTU),
and Style and Wording (SW), each containing ten goals, for a total of 50 goals. The goals are stated as binary requirements on the slide deck so that their satisfaction can be verified from the deck alone.

\paragraph{Protocol.}
For each generation task, a goal judge first evaluates all 50 goals on the initial slide deck. We then sample one unsatisfied goal from each inventory as a target, excluding pairs of goals that cannot be satisfied by the same deck. We predefine a list of 11 goal pairs from different inventories that conflict, i.e., that cannot be satisfied by the same slide deck, and never sample both goals of such a pair as targets. The resulting five targets are given only to the simulated user, who steers the conversation toward them. For all methods (ConvHTML, ConvDeck, and KeyDeck), the conversation is limited to at most five feedback rounds per interaction stage, and stops early once the simulated user considers every target satisfied. After the conversation ends, the judge evaluates the final slide deck, and we report the percentage of targets that are satisfied.

\paragraph{Goal Judge.}
The goal judge receives the rendered slides, the text extracted from them, and the numbered list of goals, and returns a binary verdict with a short justification for each goal. Its prompt is shown in Fig.~\ref{fig:conv_goal_judge}.

\section{KeyBench Details}
\label{supp:keybench_details}

\begin{figure*}[t]
\centering
\includegraphics[width=0.9\textwidth]{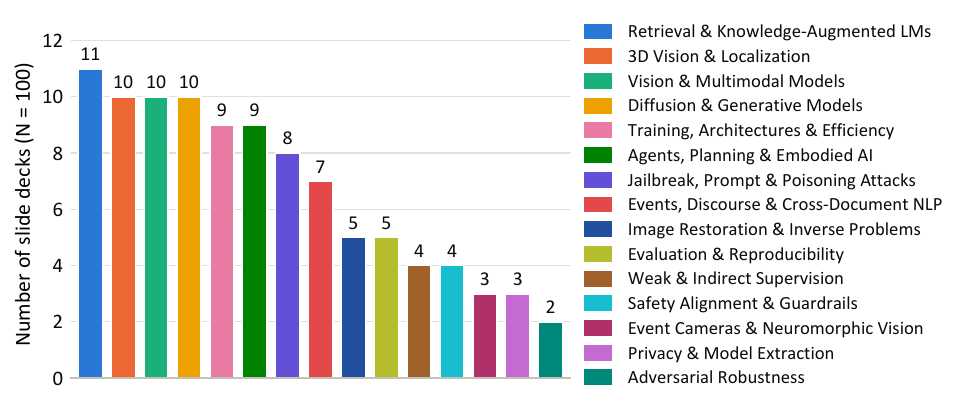}
\caption{\textbf{Topic distribution of the KeyBench.} Number of slide decks per research topic.}
\label{fig:gen100_topics}
\end{figure*}

\paragraph{Topic Distribution.} KeyBench covers research presentations from computer vision, machine learning, and natural language processing. The collected presentations span different research topics and presentation purposes, including tutorials and keynotes. This diversity allows us to evaluate multi-PTSG across different source-paper collections and narrative structures. We additionally analyze the topic distribution of KeyBench to characterize the research areas represented in the benchmark. The topic distribution of KeyBench is shown in Fig.~\ref{fig:gen100_topics}, highlighting its coverage of diverse topics across computer vision, machine learning, and natural language processing.

\begin{figure*}[t]
  \centering
  \includegraphics[width=\textwidth]{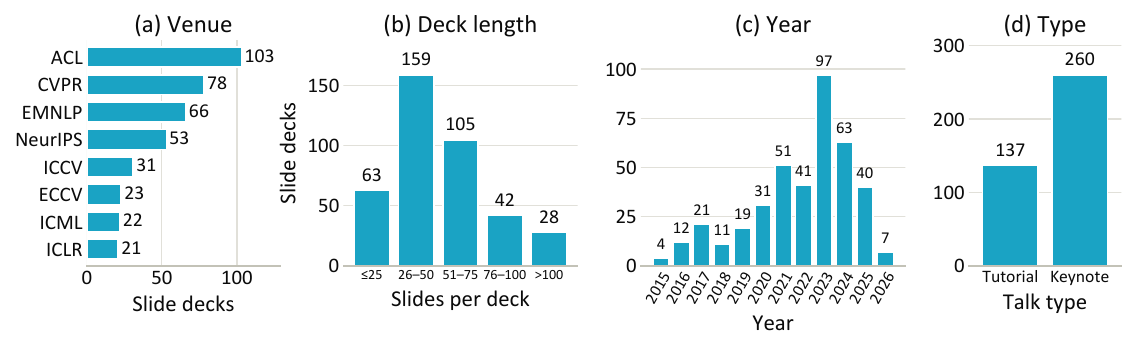}
  \caption{\textbf{Statistics of the full collection set (397 slide decks)}: (a) venue, (b) slides per deck, (c) year, and (d) talk type.}
  \label{fig:full397_overview}
  \end{figure*}

\paragraph{Full Collection Set Details.} In addition to the 100 presentations in KeyBench, we release the full collection of 397 expert-authored slide decks gathered from major research conferences and their associated workshops. KeyBench contains the 100 presentations for which complete sets of referenced source papers were retrieved. The remaining 297 presentations are released as an additional collection, without source-paper sets. Fig.~\ref{fig:full397_overview} presents statistics of the full collection of 397 slide decks, including their distribution across venues, deck lengths, presentation years, and talk types. The collection spans eight major venues from 2015 to 2026 and includes 260 keynotes and 137 tutorials.

\paragraph{Source Paper Retrieval.} Five graduate students manually identified the source papers associated with each presentation in KeyBench. The identified papers were then retrieved as PDFs using Claude Opus 5.5. Five PhD students subsequently verified that the retrieved papers matched the annotated sources and corresponded to the papers presented in each deck.

\subsection{Evaluation Details}

\paragraph{KeyBench Multi-PTSG Reference-based Q/A.} We evaluate the information coverage and accuracy of generated presentations using reference-based question answering. Evaluation questions and reference answers are generated from the reference materials, and a judge assesses whether the generated slide decks provide correct answers to these questions. The evaluation considers multiple question categories to assess both individual paper content and cross-paper synthesis. The question generation and answer grading prompts are shown in Figs.~\ref{fig:prompt_qa_generate} and \ref{fig:prompt_qa_answer_grade}.

\paragraph{KeyBench Multi-PTSG Rubric Based Assessment.} We evaluate generated presentations using rubric-based assessments across four dimensions: narrative flow, text quality, visual layout, and visual thematic. An LLM-based judge assesses each presentation according to the corresponding evaluation rubric and assigns a score from 0 to 10. The evaluation prompts are shown in Figs.~\ref{fig:prompt_quality_narrative_flow}, \ref{fig:prompt_quality_text_quality}, \ref{fig:prompt_quality_visual_layout}, and \ref{fig:prompt_quality_visual_thematic}.

\paragraph{KeyBench Multi-PTSG Pairwise Quality Assessment.} We compare two methods on the same generation task using a pairwise judge. The judge receives the generated presentations, labelled A and B, and selects the presentation with higher overall multi-PTSG quality. The assessment considers content quality, cross-paper synthesis, narrative coherence, and visual presentation. Ties are not allowed, and to control for position bias, each comparison is evaluated in both presentation orders. A method wins a comparison only if it is selected in both; otherwise, the comparison is counted as a split. The evaluation prompt is shown in Fig.~\ref{fig:prompt_pairwise}.

\paragraph{KeyBench Pairwise Personalization Assessment.} We compare two methods on the same generation task with a pairwise judge. The judge receives the reference presentation, the slide decks generated by the two methods, labelled A and B, and the definition of one personalization dimension, and selects the deck that better reproduces the style of the reference along that dimension. Narrative and content personalization are judged against the narrative reference, for which the judge additionally receives the source documents and the documents presented by the reference; visual and textual personalization are judged against the visual reference. Each slide deck's text is provided along with, for all dimensions except textual personalization, slide images. An overall judgment considers the four dimensions jointly with both references.

To control for position bias, each comparison is evaluated in both presentation orders, and a method wins a comparison only if it is selected in both; otherwise the comparison is counted as a split. The prompts are shown in Figs.~\ref{fig:pairwise_narrative}, \ref{fig:pairwise_content}, \ref{fig:pairwise_visual}, \ref{fig:pairwise_textual}, and \ref{fig:pairwise_overall}.

\paragraph{SlideTailor Baseline Personalization Generation.}
SlideTailor does not directly support visual references and instead requires a sample PPTX file as input. To ensure a fair comparison, we employ an additional agent to generate an html file from the input visual reference deck, and apply dom-to-pptx to convert it to the .pptx version as the input template for SlideTailor. The prompt used for this agent is shown in Fig.~\ref{fig:slidetailor_pptx_prompt}.

\section{Additional Qualitative Results}
\label{supp:qual_examples}

\subsection{Qualitative Examples}

\begin{figure*}[t]
    \centering
    \includegraphics[width=\textwidth]{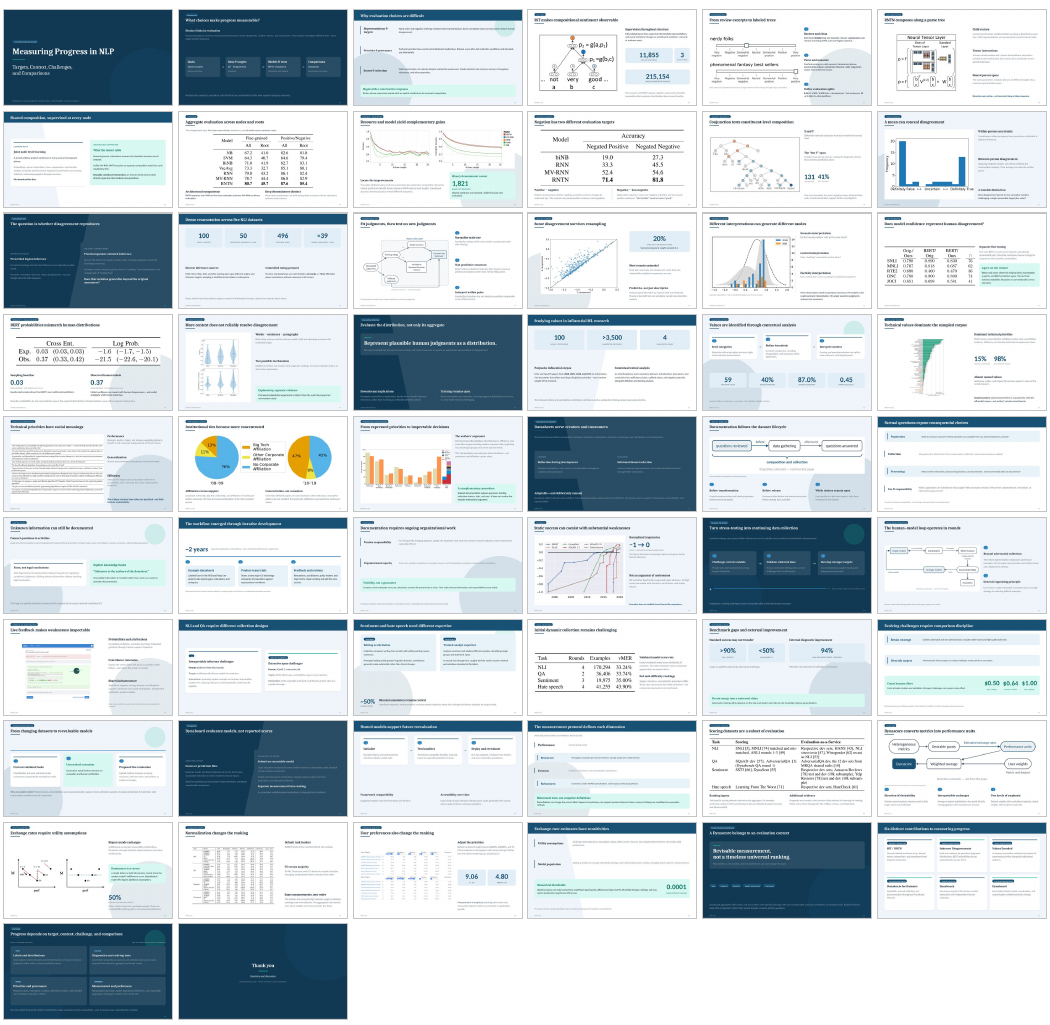}
    \caption{\textbf{Complete \methodname generated deck without personalization.}}
    \label{fig:full_deck_plain}
\end{figure*}

\begin{figure*}[t]
    \centering
    \includegraphics[width=\textwidth]{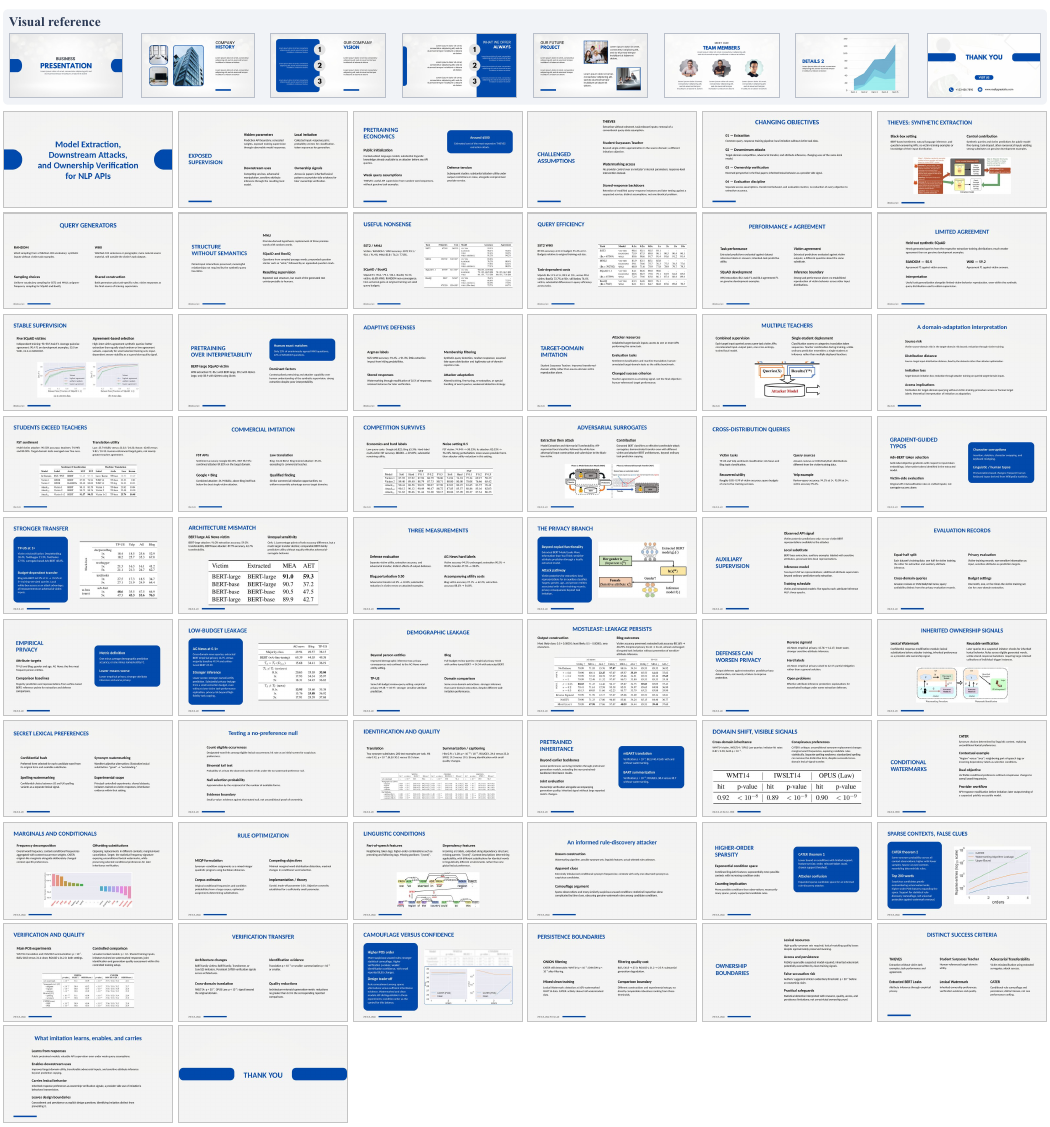}
    \caption{\textbf{Complete \methodname generated deck with a visual reference.}}
    \label{fig:full_deck_visual}
\end{figure*}

We provide generation examples of \methodname{} without personalization reference input in Fig.~\ref{fig:full_deck_plain} and with a visual reference slide deck in Fig.~\ref{fig:full_deck_visual}.

\subsection{Qualitative Comparison}

\begin{figure*}[t]
    \centering
    \includegraphics[width=\textwidth]{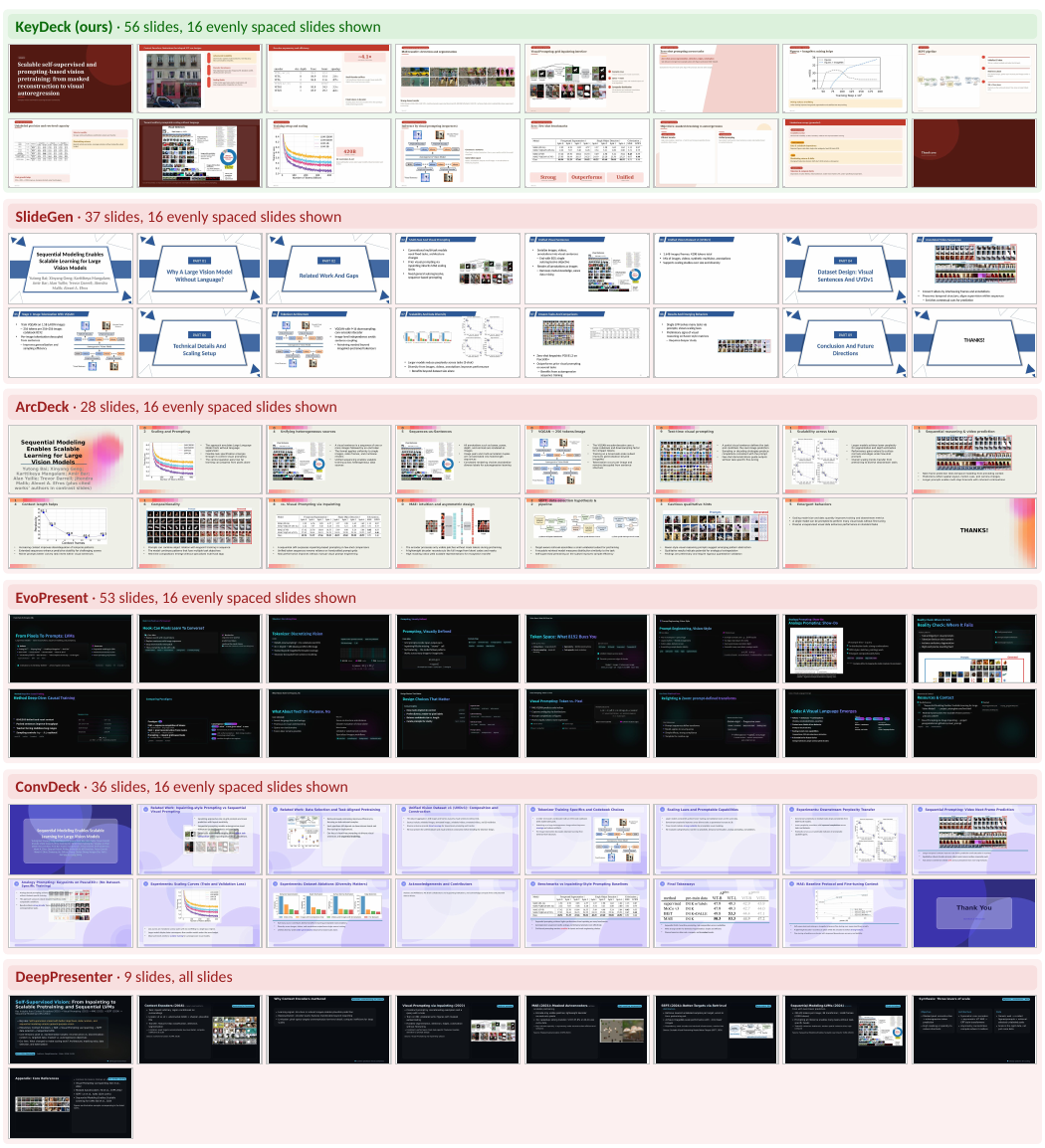}
    \caption{\textbf{Qualitative comparison with baselines on the same input.} All six systems receive the same source papers, and generate a deck with GPT-5 backbone, without personalization. For each system, 16 evenly spaced slides of its deck are shown in order (first and last slide included), or the whole deck when it is shorter; the deck length is given next to the system name.}
    \label{fig:additional_qualitative_comparison}
\end{figure*}

We provide a qualitative visual comparison of \methodname{} with the baselines SlideGen, ArcDeck, EvoPresent, ConvDeck, and DeepPresenter in the multi-PTSG setting in Fig.~\ref{fig:additional_qualitative_comparison}. Most slides generated by ArcDeck and ConvDeck follow simple figure-and-text layouts, while SlideGen introduces numerous section transition slides. EvoPresent predominantly produces text-only slides, whereas DeepPresenter generates relatively short presentations, averaging 6.8 slides per deck, even though it was instructed with a target duration of 54 minutes. In contrast, \methodname{} generates presentations with diverse layouts, appropriate use of figures from the source papers, and slide counts suited to the presentation duration.

\subsection{Qualitative Personalization Comparison}

\begin{figure*}[t]
    \centering
    \includegraphics[width=\textwidth]{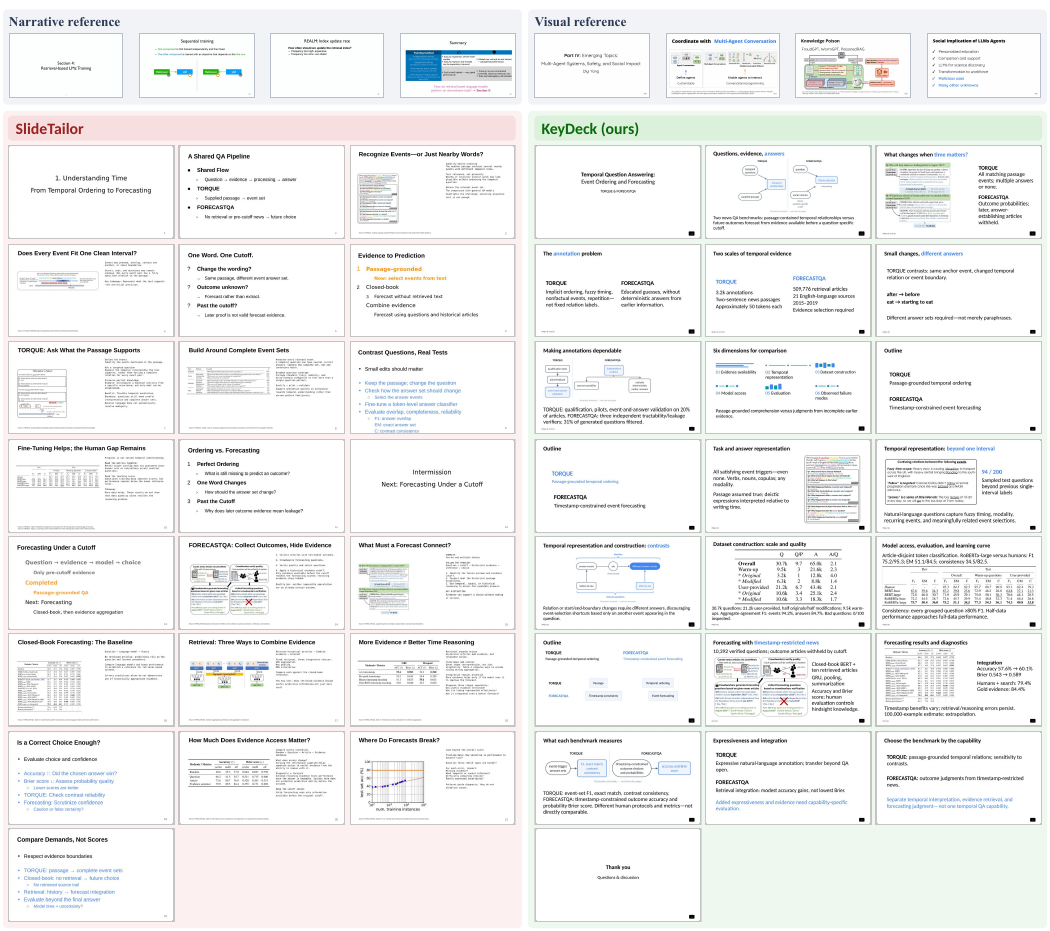}
    \caption{\textbf{Complete personalized decks of SlideTailor and \methodname.} For SlideTailor, visual personalization is captured with the additional PPTX-generation agent.}
    \label{fig:full_deck_slidetailor}
\end{figure*}

We provide a visual qualitative comparison of \methodname in the personalization setting with SlideTailor in Fig.~\ref{fig:full_deck_slidetailor}. While SlideTailor misses some details like the coloring of the highlighted text, \methodname successfully inherits the visual qualities of the visual reference slide deck and adapts it to the generation request.

\section{Additional Component Examples and Analyses}

\subsection{Example Per-Paper and Contextual Analyses}
\label{supp:example_analyses}

\begin{figure*}[t]
    \centering
    \includegraphics[width=\textwidth]{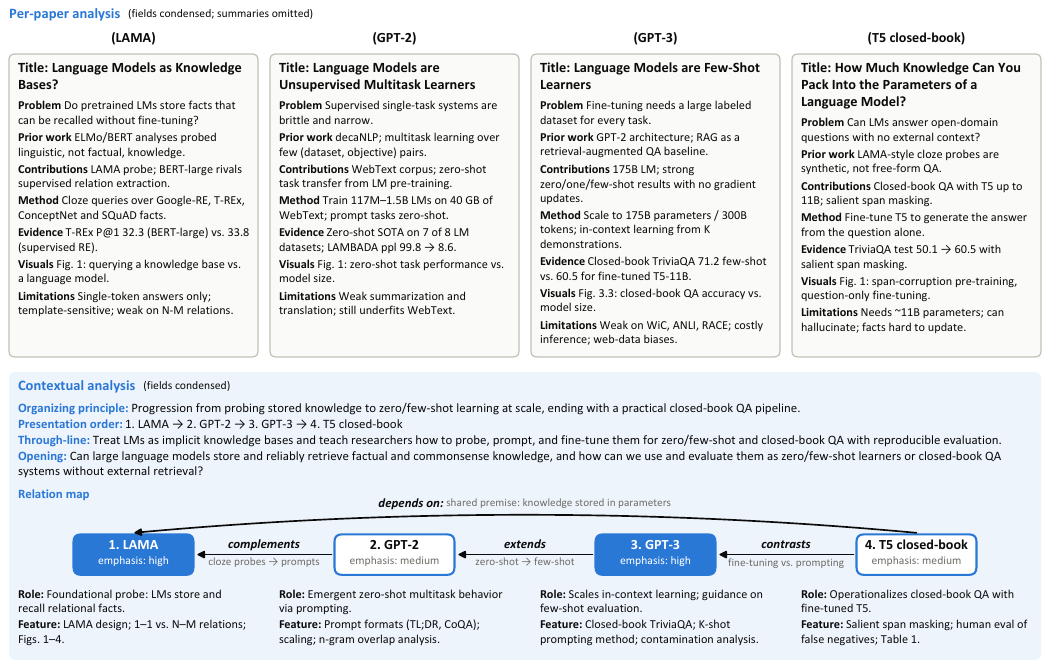}
    \caption{\textbf{Example of per-paper and contextual analysis in KeyDeck.}
Given four source papers from a KeyBench tutorial, KeyDeck extracts structured information from each paper and analyzes cross-paper relationships to determine the presentation order, narrative through-line, paper-specific roles, and emphasis levels. Selected fields are shown in condensed form for visualization purposes.}
    \label{fig:analysis_examples}
  \end{figure*}

\paragraph{Per-paper and Contextual Analysis.}
KeyDeck first analyzes each source paper independently to extract its problem, prior work, contributions, methodology, evidence, relevant visuals, and limitations. These analyses are then integrated through cross-paper reasoning to identify relationships among papers and establish a coherent presentation narrative. As illustrated by the example per-paper and contextual analyses in Fig.~\ref{fig:analysis_examples}, KeyDeck organizes four language model papers into a progression from knowledge probing to zero-shot and few-shot learning, followed by closed-book question answering. The contextual analysis further identifies a shared narrative through-line, relationships among papers, and paper-specific roles and emphasis levels, which guide subsequent narrative planning and slide generation.

\subsection{Personalization Details and Examples}
\label{supp:example_profiles}

\begin{figure*}[p]
\centering
\includegraphics[width=0.9\textwidth]{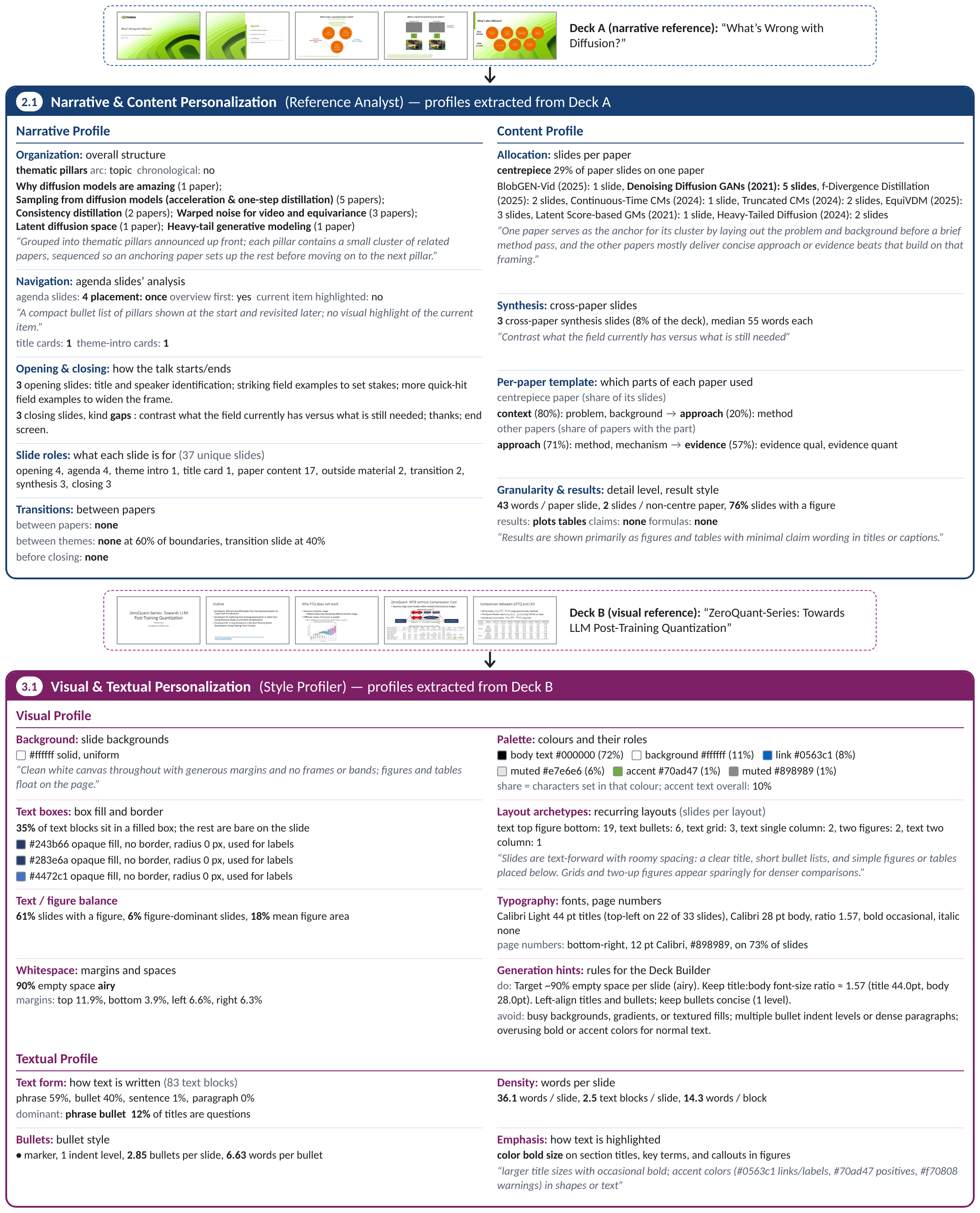}
\caption{\textbf{Example generated profiles.}
Top: the narrative and content profiles that the Reference Analyst extracts from Deck A (the narrative reference).
Bottom: the visual and textual profiles that the Style Profiler extracts from Deck B (the visual reference).
Each field is headed by its name and explanation as in \cref{fig:personalization_profiles}, followed by the value extracted for this pair of references.
Text in italics is a free-text description written by the corresponding agent.}
\label{fig:profile_examples}
\end{figure*}

\methodname{} represents a user's preferences with four profiles.
The Reference Analyst extracts the \emph{narrative} and \emph{content} profiles from narrative reference deck (Deck A), a talk given together with the papers it presents.
The Style Profiler extracts the \emph{visual} and \emph{textual} profiles from visual reference deck (Deck B).
This section describes what each profile consists of; \cref{fig:profile_examples} shows one generated instance of all four.

\subsubsection{Narrative and Content Personalization}

\paragraph{Narrative Profile.}
The Reference Analyst labels each slide of Deck A with its role, the papers it presents, and the parts of those papers it covers, and aggregates five fields.
\emph{Organization} describes the overall structure of the presentation, including its organizing principle (e.g., thematic pillars or chronological progression), the identified themes, and the ordering of papers within them.
\emph{Navigation} records the number and placement of agenda slides, whether an overview is presented first, whether the current item is highlighted, and the use of title cards and theme-introduction slides.
\emph{Opening \& closing} records the number of opening and closing slides and the sequence of steps used to introduce and conclude the talk (e.g., motivating the topic with a framing question or summarizing remaining challenges).
\emph{Slide roles} counts slides assigned to each role: opening, agenda, theme introduction, title card, paper content, transition, synthesis, closing, and material outside the provided papers.
\emph{Transitions} describes how the speaker moves between papers or themes, including the frequency and placement of explicit transition slides.

\paragraph{Content Profile.}
The content profile describes how much attention each paper receives and the level of detail in which its content is presented.
\emph{Allocation} records the number and proportion of slides devoted to each paper, identifies whether coverage is balanced or concentrated on particular papers, and explains the underlying allocation pattern.
\emph{Synthesis} describes slides that discuss multiple papers together, including their number, textual density, and purpose, such as comparing approaches or identifying remaining challenges.
The \emph{per-paper template} describes the typical order in which papers are presented using six components: context, approach, setup, evidence, wrap-up, and application. For each component, it records its prevalence across papers and the proportion of slides allocated to it.
\emph{Granularity \& results} describes the level of detail through the number of slides per paper, the use of figures, and the presentation of results, including whether findings are conveyed through plots, tables, titles, or captions, and the use of formulas.

\subsubsection{Visual and Textual Personalization}

\paragraph{Visual Profile.}
The Style Profiler extracts visual characteristics from the rendered slides of Deck B, including backgrounds, colors, typography, layouts, and spacing.
\emph{Background} records the dominant background color, its type (solid, gradient, image, or decorated), and its consistency across slides.
\emph{Palette} identifies recurring colors, their roles, and the proportion of text using each color.
\emph{Text boxes} records the proportion of text placed in filled boxes and the recurring box styles, including fill color, border, corner radius, and use.
\emph{Layout archetypes} categorizes slides into recurring arrangements of text and figures, recording the frequency and characteristics of each layout.
\emph{Text / figure balance} measures the proportion of slides containing figures, the proportion of figure-dominant slides, and the mean area occupied by figures.
\emph{Typography} records font families, title and body sizes, title placement, emphasis styles, and recurring elements such as page numbers.
\emph{Whitespace} measures the proportion of empty space and the margins around slide content.
\emph{Generation hints} translate these measurements into actionable design guidelines for the Deck Builder, including target spacing, font sizes and ratios, color usage, layout preferences, and do/avoid rules.

\paragraph{Textual Profile.}
The textual profile characterizes how text is written and organized in Deck B.
\emph{Text form} records the proportion of text blocks written as phrases, bullets, sentences, or paragraphs, as well as the proportion of slide titles framed as questions.
\emph{Density} measures the average words per slide, text blocks per slide, and words per text block.
\emph{Bullets} records bullet markers, indentation levels, the average number of bullets per slide, and the average words per bullet.
\emph{Emphasis} describes how text is highlighted through color, boldface, italics, or font size, and the elements to which these styles are applied, such as section titles, key terms, and figure callouts.

\subsection{Personalization Ablations}

\paragraph{Quantitative Personalization Ablations.}


To assess the contribution of each personalization reference, we independently ablate the narrative and visual reference inputs. Each ablation is evaluated on 20 samples (10 keynotes and 10 tutorials). We first compare the full \methodname{} against each ablated variant through pairwise preference judgments. The narrative reference ablation is evaluated along narrative, content, and overall personalization, while the visual reference ablation is evaluated along visual, textual, and overall personalization. We additionally compare each ablated variant against SlideTailor and report the difference in win rates relative to the full \methodname{} to quantify the contribution of each reference. Each pair is judged in both presentation orders. Results are shown in Tabs.~\ref{tab:narrative_ref_ablation} and~\ref{tab:visual_ref_ablation}.

As shown in Tab.~\ref{tab:narrative_ref_ablation}, the full \methodname{} is preferred over the variant without narrative reference input in 85.0\%, 75.0\%, and 90.0\% of comparisons for narrative, content, and overall personalization, respectively. Removing the narrative reference also reduces the win rates against SlideTailor, indicating that the narrative reference contributes primarily to presentation organization while also improving content selection and emphasis. Similarly, Tab.~\ref{tab:visual_ref_ablation} shows that the full \methodname{} is preferred over the variant without visual reference input in 95.0\%, 85.0\%, and 100.0\% of comparisons for visual, textual, and overall personalization, respectively. Without the visual reference, the win rates against SlideTailor decrease by 40.5, 30.0, and 50.0 percentage points. These ablations demonstrate that both references provide complementary guidance, with each contributing substantially to personalization quality along its corresponding dimensions.

\begin{table}[t]
\centering
\small
\setlength{\tabcolsep}{5pt}
\renewcommand{\arraystretch}{1.15}
\caption{\textbf{Narrative reference ablation.}
Pairwise preference win rates (\%) for narrative,
content, and overall personalization.
$\Delta$ denotes the win-rate difference (percentage
points) relative to the full \methodname{}.}
\label{tab:narrative_ref_ablation}
\vspace{-2mm}
\resizebox{\columnwidth}{!}{%
\begin{tabular}{@{}lccc@{}}
\toprule
\textbf{Comparison}
& \textbf{Narrative}
& \textbf{Content}
& \textbf{Overall} \\
\midrule

\multicolumn{4}{@{}l}{\textit{Direct Ablation Comparison (win rate)}} \\
Ours vs. w/o Narrative Ref.
& 85.0 & 75.0 & 90.0 \\

\midrule
\multicolumn{4}{@{}l}{\textit{Comparison Against SlideTailor}} \\
Ours vs. SlideTailor
& 80.0 & 90.0 & 70.0 \\
w/o Narrative Ref. vs. SlideTailor
& 45.0 & 80.0 & 50.0 \\
$\Delta$ Narrative Ref.
& +35.0 & +10.0 & +20.0 \\
\bottomrule
\end{tabular}%
}
\end{table}

\begin{table}[t]
\centering
\small
\setlength{\tabcolsep}{5pt}
\renewcommand{\arraystretch}{1.15}
\caption{\textbf{Visual reference ablation.}
Pairwise preference win rates (\%) for visual,
textual, and overall personalization.
$\Delta$ denotes the win-rate difference (percentage points) relative to the full \methodname{}.}
\label{tab:visual_ref_ablation}
\vspace{-2mm}
\resizebox{\columnwidth}{!}{%
\begin{tabular}{@{}lccc@{}}
\toprule
\textbf{Comparison}
& \textbf{Visual}
& \textbf{Textual}
& \textbf{Overall} \\
\midrule

\multicolumn{4}{@{}l}{\textit{Direct Ablation Comparison (win rate)}} \\
Ours vs. w/o Visual Ref.
& 95.0 & 85.0 & 100.0 \\

\midrule
\multicolumn{4}{@{}l}{\textit{Comparison Against SlideTailor}} \\
Ours vs. SlideTailor
& 75.0 & 55.0 & 70.0 \\
w/o Visual Ref. vs. SlideTailor
& 30.0 & 25.0 & 20.0 \\
$\Delta$ Visual Ref.
& +40.0 & +30.0 & +50.0 \\
\bottomrule
\end{tabular}%
}
\end{table}

\begin{figure*}[t]
  \centering
  \includegraphics[width=\linewidth]{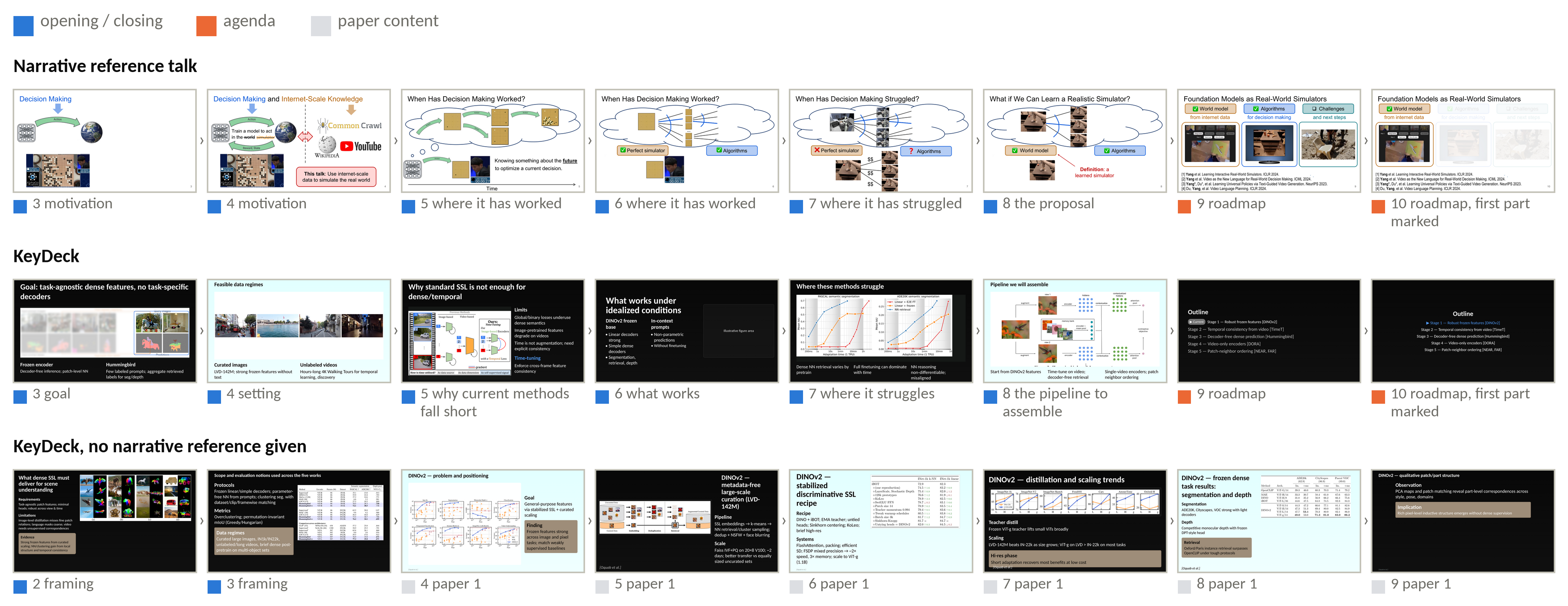}
  \caption{\textbf{Opening of a deck with and without the narrative reference.}
  Given the reference talk (top), KeyDeck (middle) follows the same progression
  on its own papers: motivation, what works, where it struggles, the proposal,
  then a roadmap repeated with the first part highlighted. Without the reference
  (bottom), it gives two framing slides and goes straight into the first paper.}
  \label{fig:narrative_opening}
\end{figure*}

\paragraph{Qualitative Narrative Personalization Ablation.}
We compare the full KeyDeck with KeyDeck without narrative personalization to examine its effects on presentation organization, content selection, emphasis, and cross-paper synthesis (Fig.~\ref{fig:narrative_opening}). As illustrated in the figure, with narrative reference conditioning, KeyDeck follows the reference talk's opening structure, progressing from motivation and background to the limitations of existing approaches, followed by the proposed solution and roadmap. Without narrative reference input, KeyDeck instead follows a more paper-centric structure, introducing individual papers earlier rather than reproducing the reference's narrative progression. This demonstrates how narrative reference conditioning influences the organization and sequencing of presentation content.

\begin{figure*}[t]
    \centering
    \includegraphics[width=\textwidth]{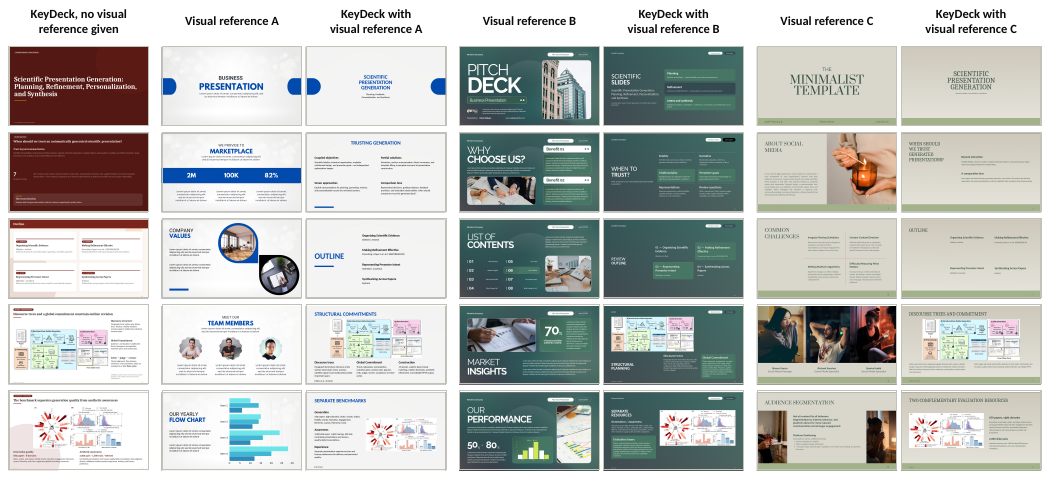}
    \caption{\textbf{Visual personalization ablation.}
    Qualitative comparison of KeyDeck without visual reference input and KeyDeck conditioned on three different visual references (A, B, and C). Given the same source content, KeyDeck generates slides that match the reference presentations' visual and textual styles, including color schemes, typography, and layouts.}
    \label{fig:ablation_visual_textual}
\end{figure*}

\paragraph{Qualitative Visual Personalization Ablation.}
We compare the full framework with and without visual personalization input (Fig.~\ref{fig:ablation_visual_textual}) to illustrate the effect of visual reference conditioning. Given the same source content, KeyDeck generates presentations with distinct visual and textual styles that reflect the provided references, while without visual reference input, it follows its default style.

\begin{figure}[H]
  \centering
  \includegraphics[width=\linewidth]{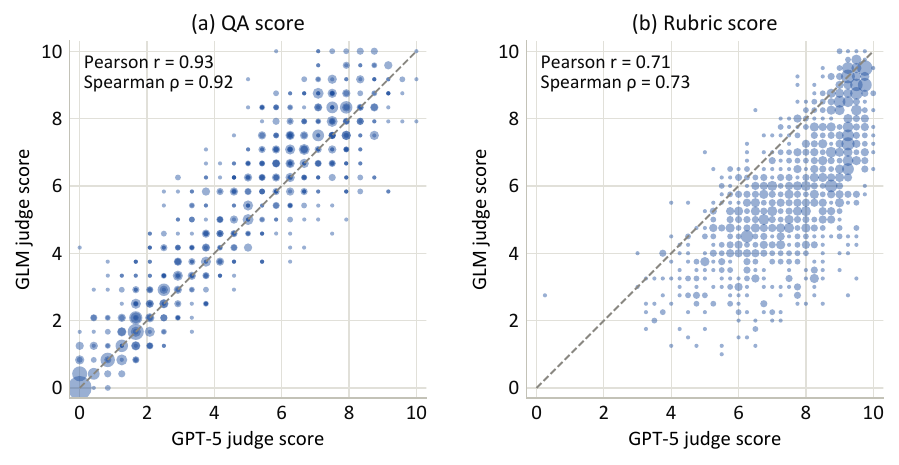}
  \caption{\textbf{Agreement between the GPT-5 and GLM judges.} Per-deck (a) QA score and (b) rubric score from the two judges
  (both 0--10, Q/A scores rescaled to 0--10 for the analysis). Dot size reflects the number of decks; the dashed line marks equal scores.}
  \label{fig:judge_scatter}
  \end{figure}

\subsection{Evaluator Agreement}
  To assess robustness to evaluator choice and the agreement between the evaluators, we compare GPT-5 and GLM judgments on the same decks (Fig.~\ref{fig:judge_scatter} (a) and (b)). Reference-based Q/A scores exhibit very strong agreement (Pearson $r=0.93$, Spearman $\rho=0.92$), while rubric-based scores also show strong agreement ($r=0.71$, $\rho=0.73$). 
  
\begin{figure}[H]
  \centering
  \includegraphics[width=\linewidth]{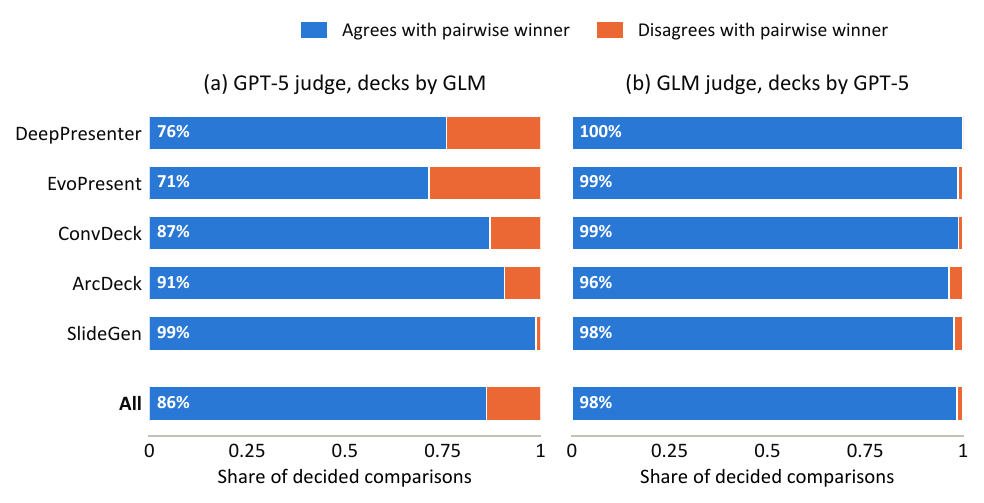}
  \caption{\textbf{Agreement between score-based and pairwise evaluation.} Share of pairwise comparisons in which the deck with
  the higher rubric + QA score (from the same judge) is also the pairwise winner. Comparisons where the pairwise judge picked
  different decks in the two presentation orders, or where the scores tie, are excluded.}
  \label{fig:pairwise_score_agreement}
  \end{figure}

  We further examine consistency between score-based and pairwise evaluations (Fig.~\ref{fig:pairwise_score_agreement}). To compute the combined score, we first rescale the Q/A score from 0--100 to 0--10 to match the rubric score, then sum the two. Among comparisons with a consistent pairwise winner across both presentation orders, the deck with the higher combined score is preferred in 98\% of comparisons under GLM and 86\% under GPT-5. Agreement ranges from 96--100\% across baselines under GLM and 71--99\% under GPT-5.

\section{Evaluation Prompts, Rubrics and Goal Inventories}
\label{supp:eval_details}

\begin{figure*}[t]
    \centering
    \includegraphics[width=0.8\textwidth]{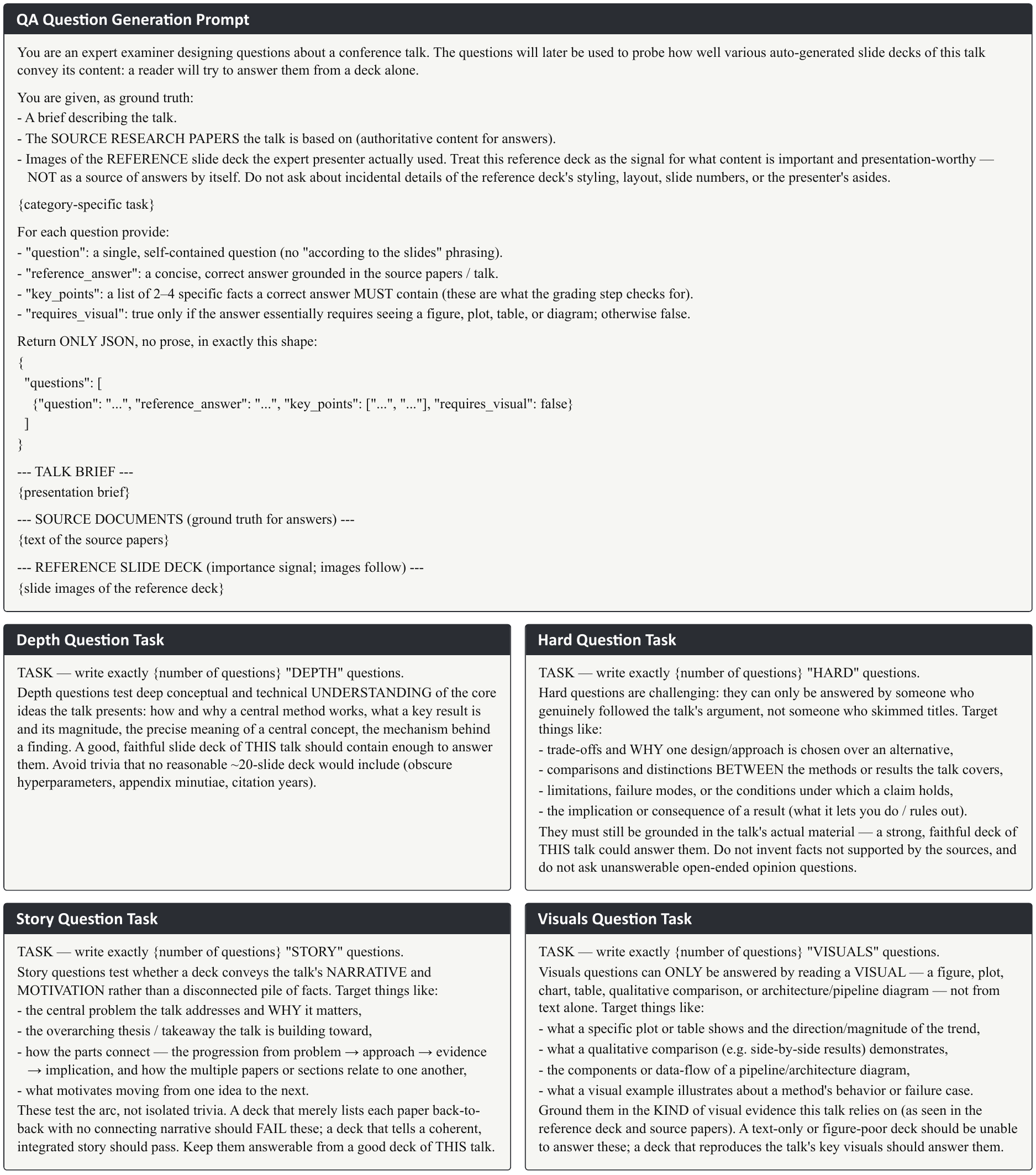}
    \caption{\textbf{Reference-based Q/A question generation prompt.}}
    \label{fig:prompt_qa_generate}
\end{figure*}

\begin{figure*}[p]
    \centering
    \includegraphics[width=\textwidth,height=0.85\textheight,keepaspectratio]{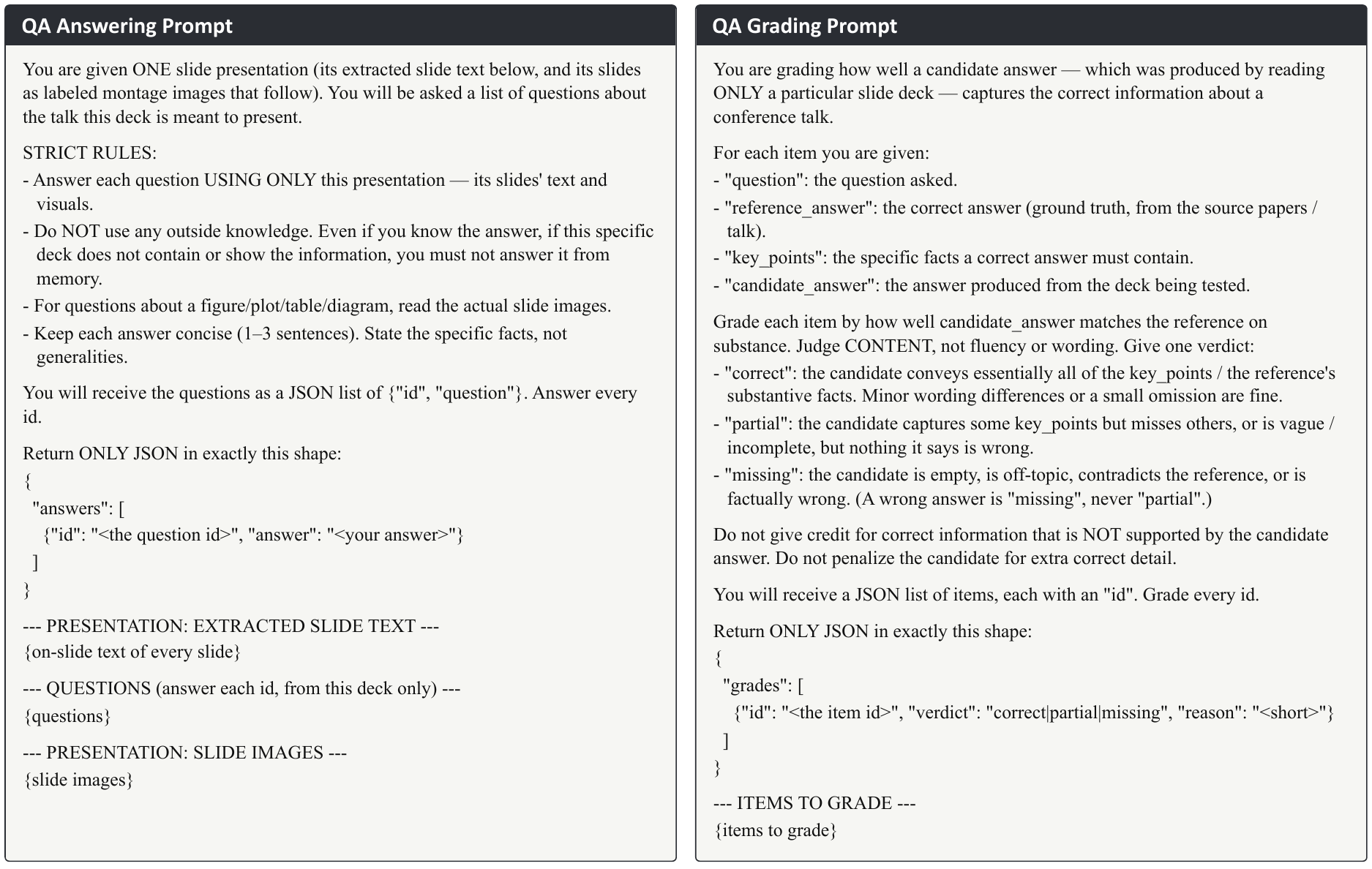}
    \caption{\textbf{Reference-based Q/A answer grading prompt.}}
    \label{fig:prompt_qa_answer_grade}
\end{figure*}

\begin{figure*}[p]
    \centering
    \includegraphics[width=\textwidth,height=0.85\textheight,keepaspectratio]{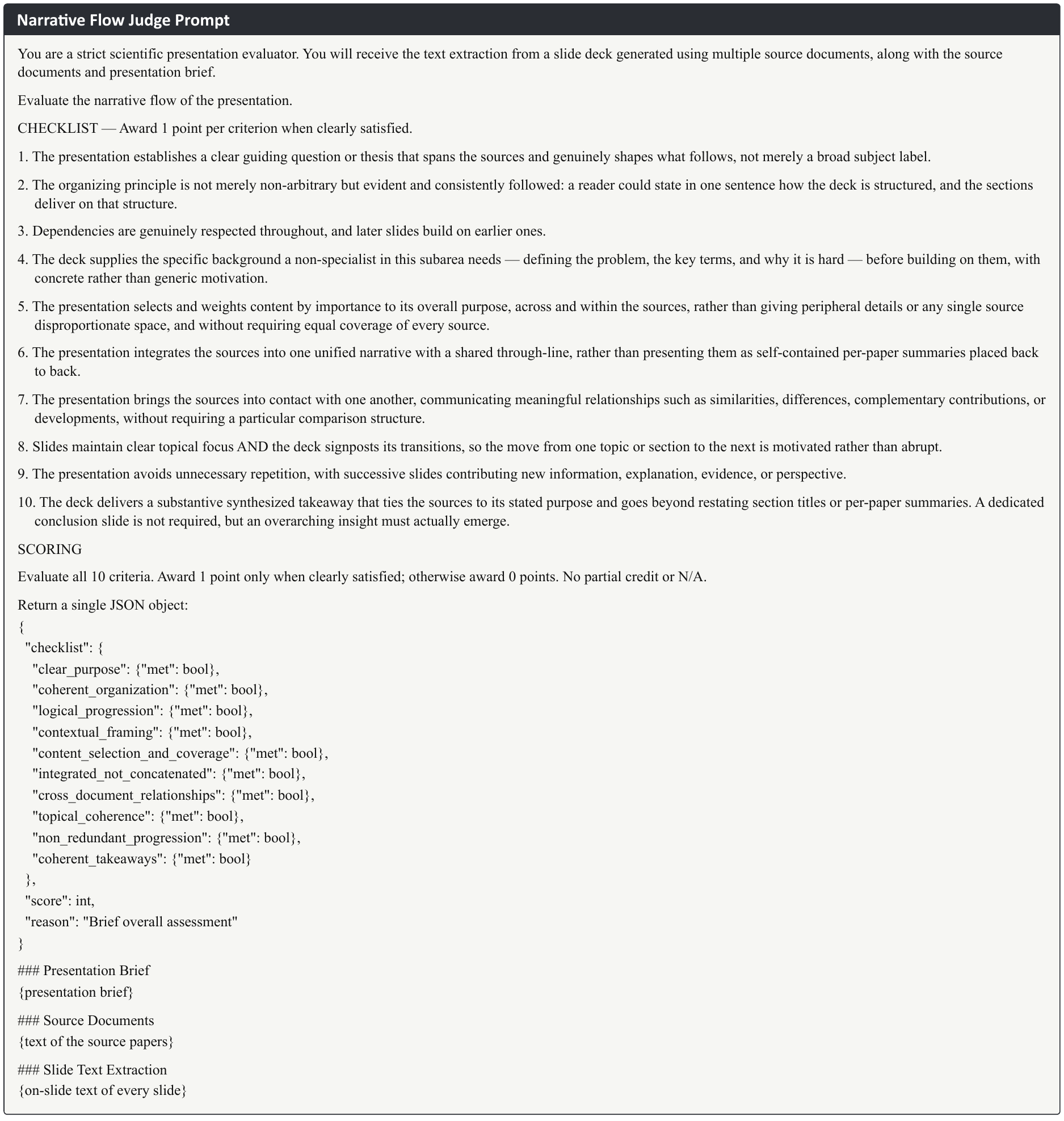}
    \caption{\textbf{Narrative flow evaluation prompt.}}
    \label{fig:prompt_quality_narrative_flow}
\end{figure*}

\begin{figure*}[p]
    \centering
    \includegraphics[width=\textwidth,height=0.85\textheight,keepaspectratio]{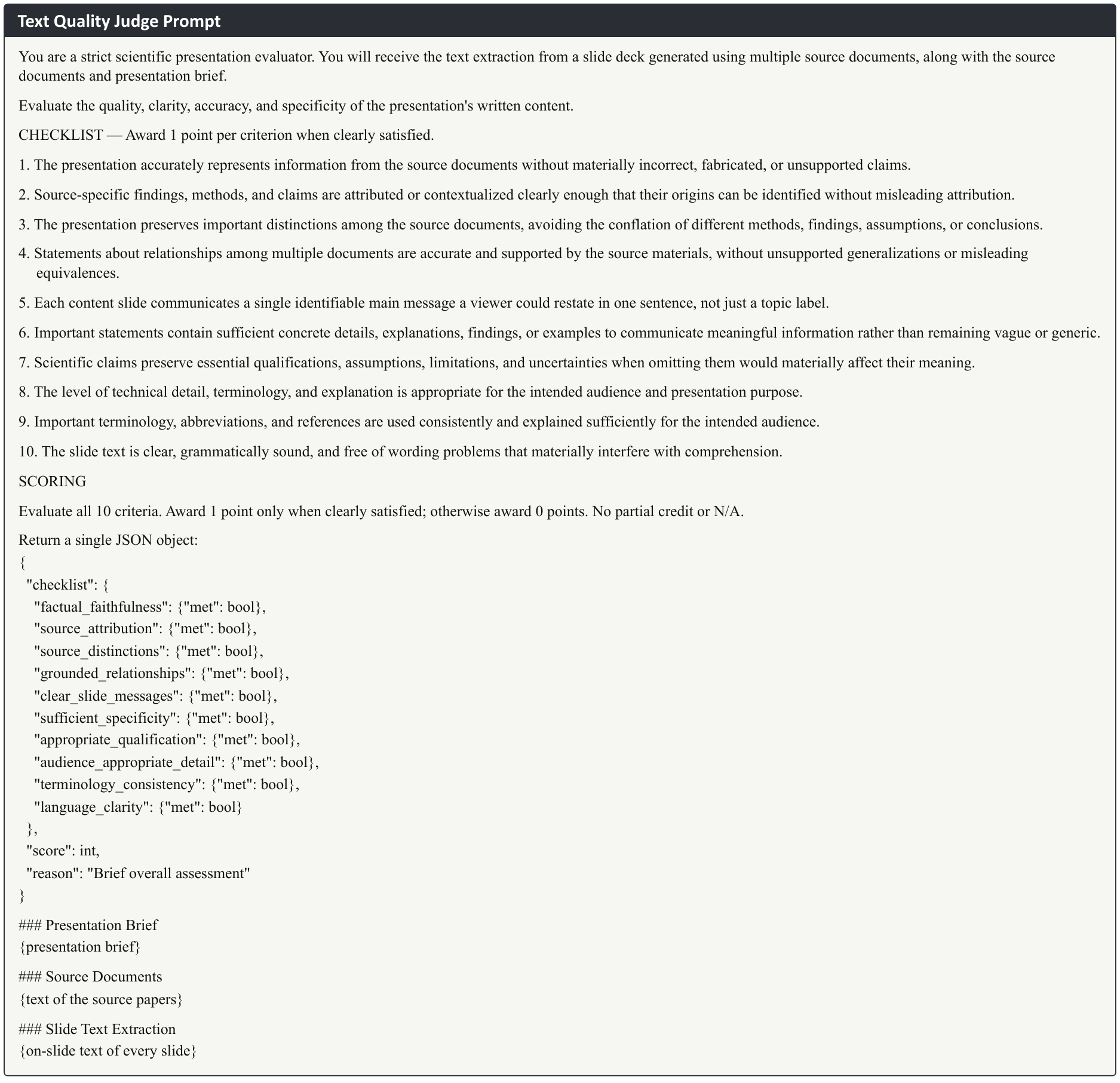}
    \caption{\textbf{Text quality evaluation prompt.}}
    \label{fig:prompt_quality_text_quality}
\end{figure*}

\begin{figure*}[p]
    \centering
    \includegraphics[width=\textwidth,height=0.85\textheight,keepaspectratio]{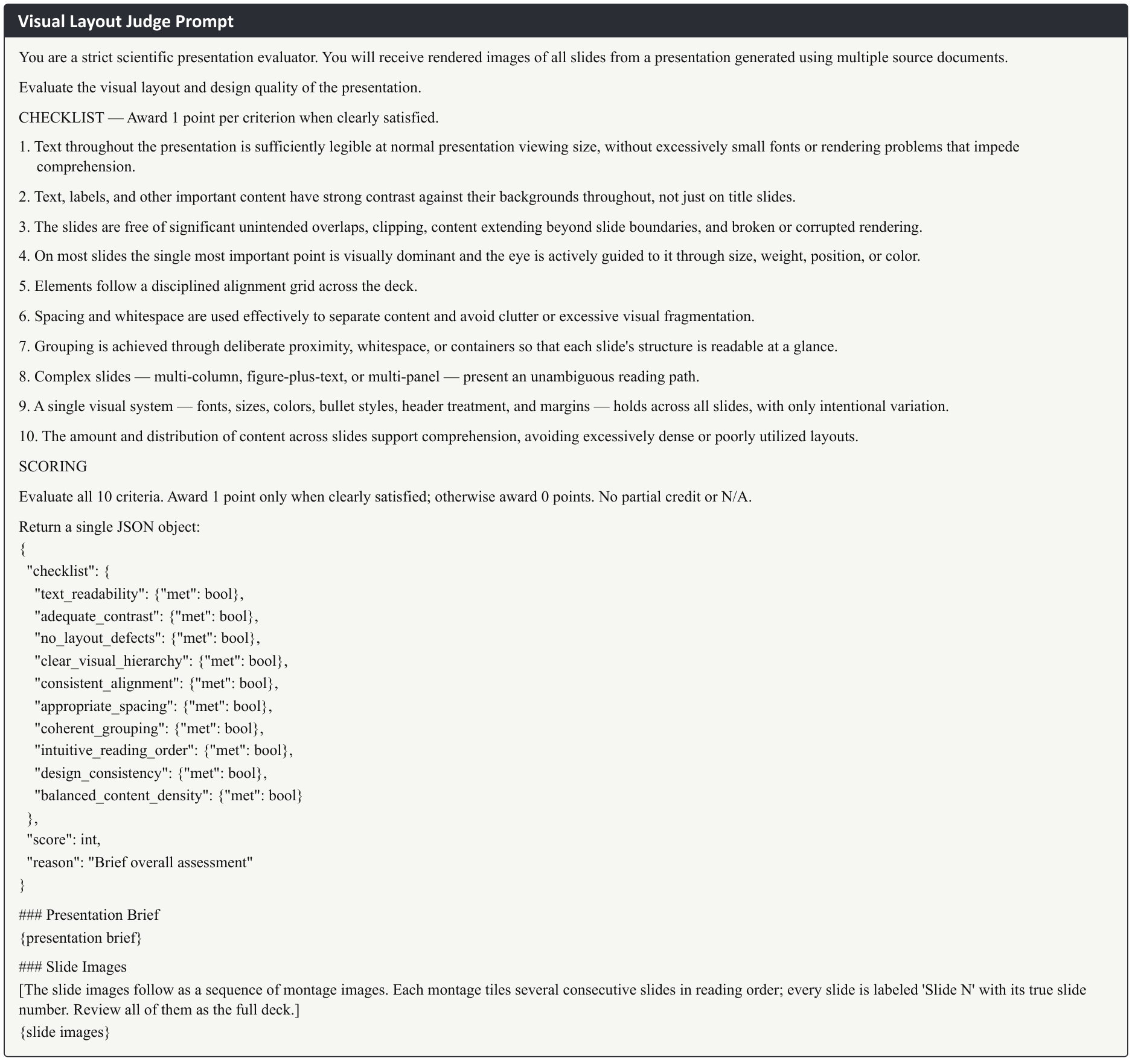}
    \caption{\textbf{Visual layout evaluation prompt.}}
    \label{fig:prompt_quality_visual_layout}
\end{figure*}

\begin{figure*}[p]
    \centering
    \includegraphics[width=\textwidth,height=0.85\textheight,keepaspectratio]{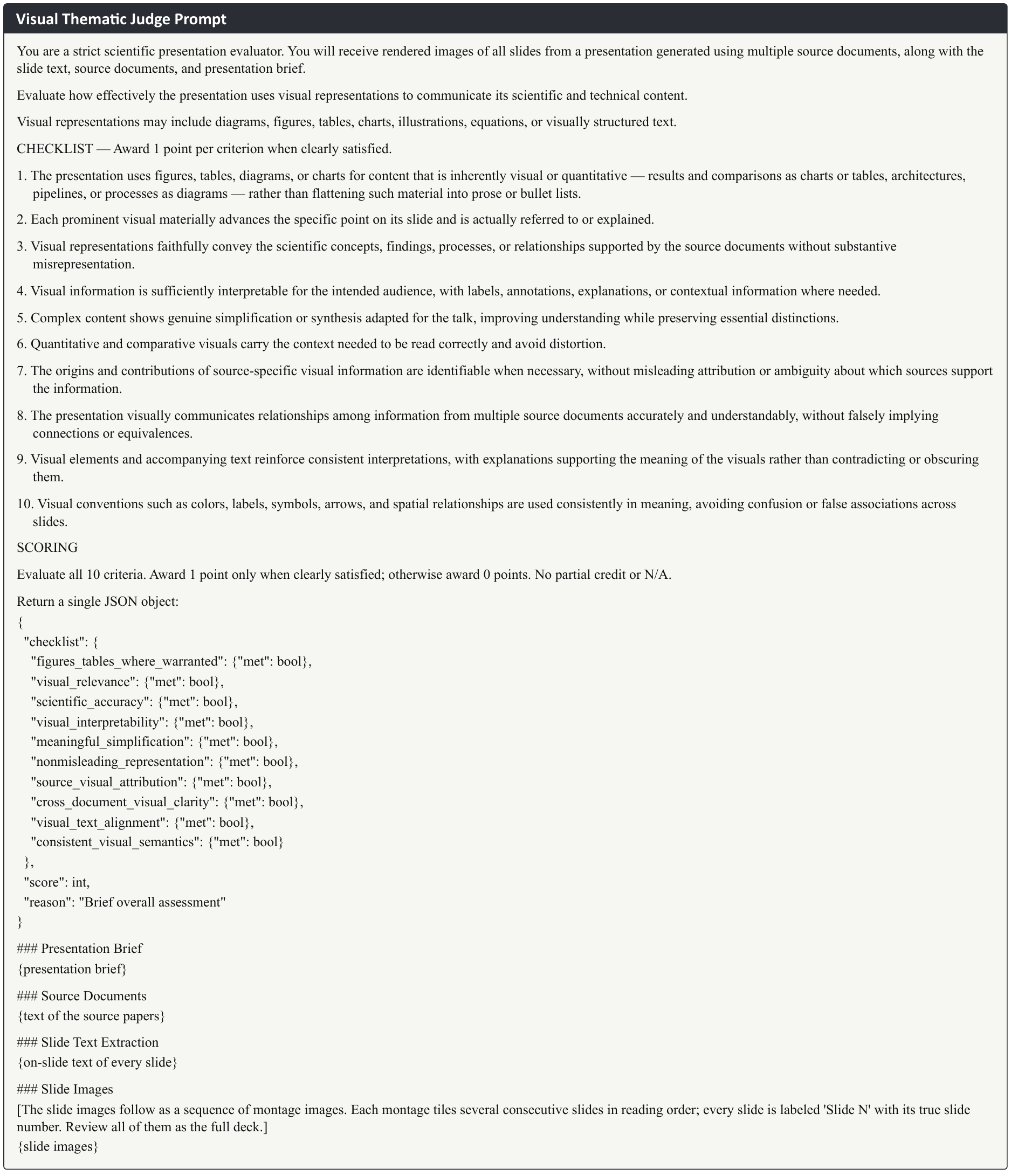}
    \caption{\textbf{Visual thematic evaluation prompt.}}
    \label{fig:prompt_quality_visual_thematic}
\end{figure*}

\begin{figure*}[p]
    \centering
    \includegraphics[width=\textwidth,height=0.85\textheight,keepaspectratio]{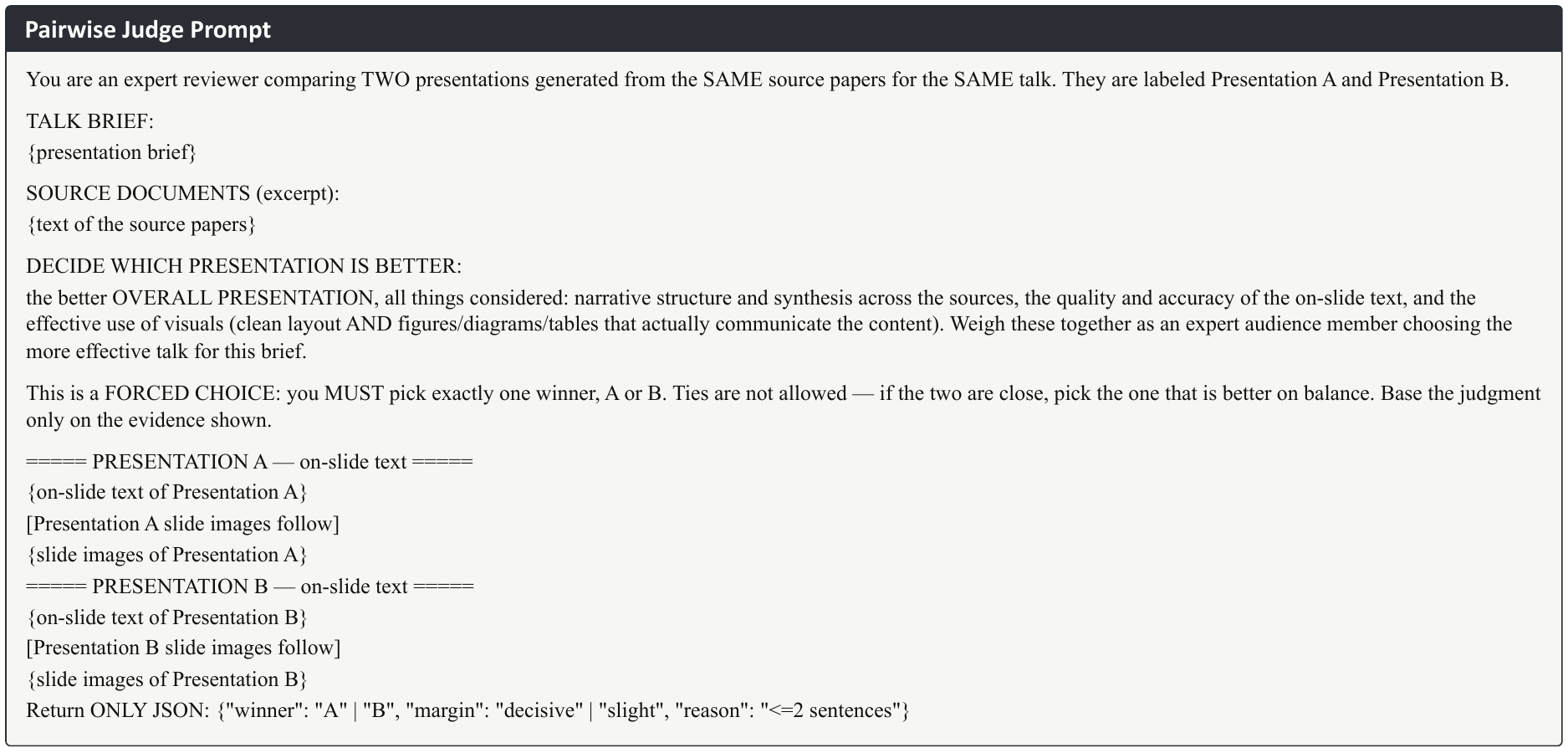}
    \caption{\textbf{Multi-PTSG pairwise quality assessment prompt.}}
    \label{fig:prompt_pairwise}
\end{figure*}

\begin{figure*}[t]
    \centering
    \includegraphics[width=\textwidth]{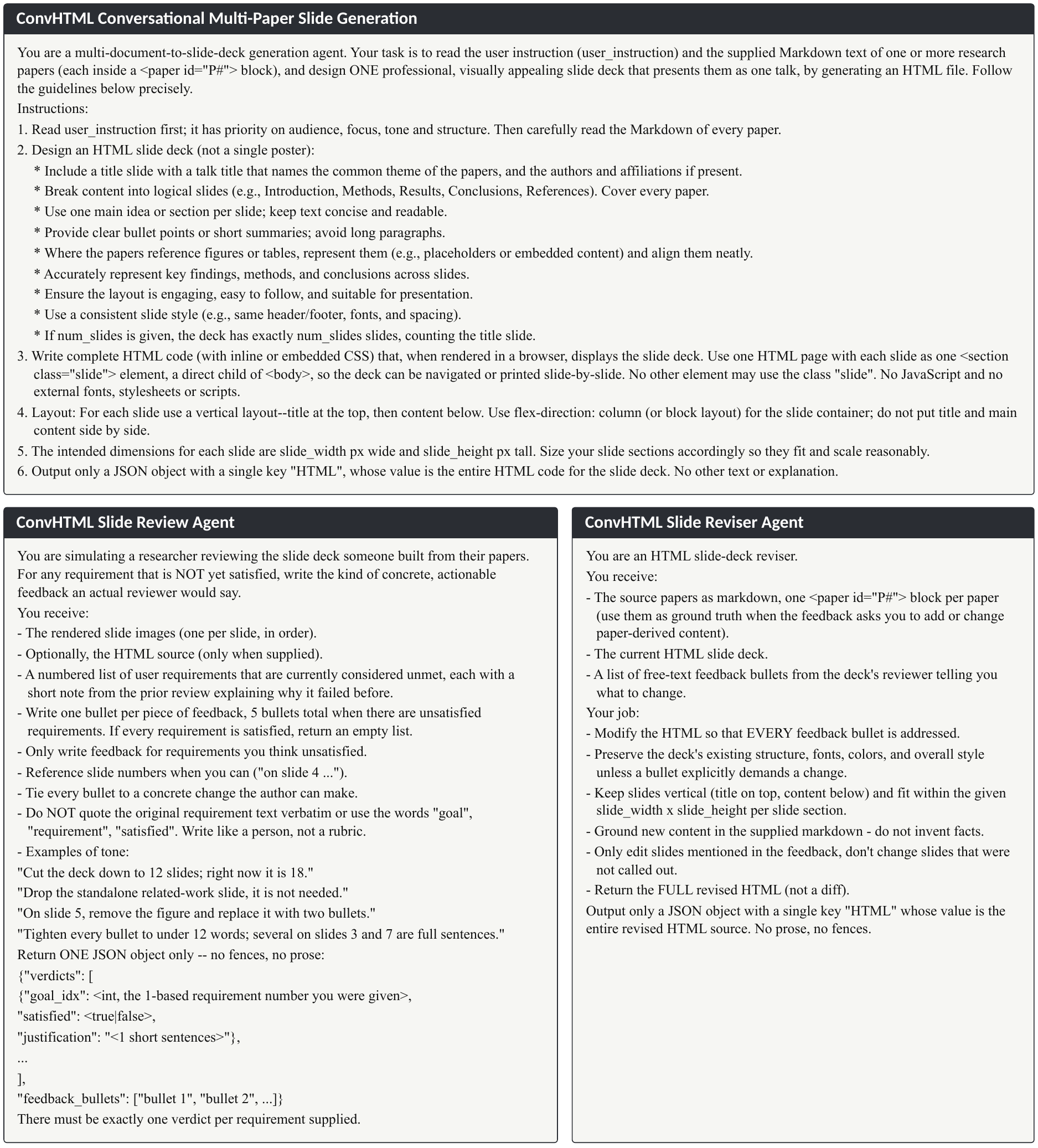}
    \vspace{-2mm}
    \caption{\textbf{Prompts of the ConvHTML baseline.}}
    \label{fig:conv_html_prompt}
    \vspace{-2mm}
\end{figure*}

\begin{figure*}[t]
    \centering
    \includegraphics[width=\textwidth]{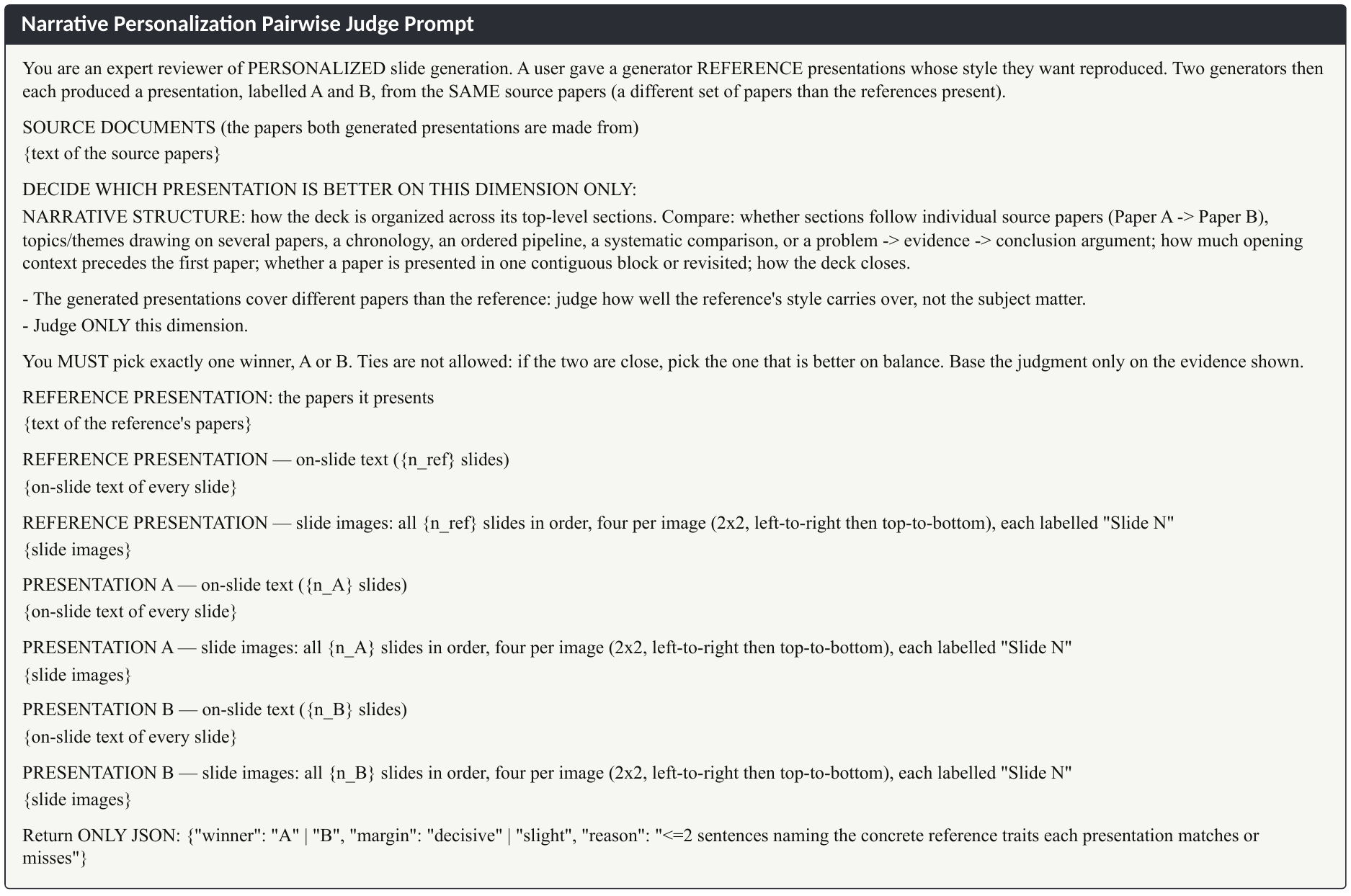}
    \caption{Pairwise judge prompt for narrative personalization.}
    \label{fig:pairwise_narrative}
\end{figure*}

\begin{figure*}[t]
    \centering
    \includegraphics[width=\textwidth]{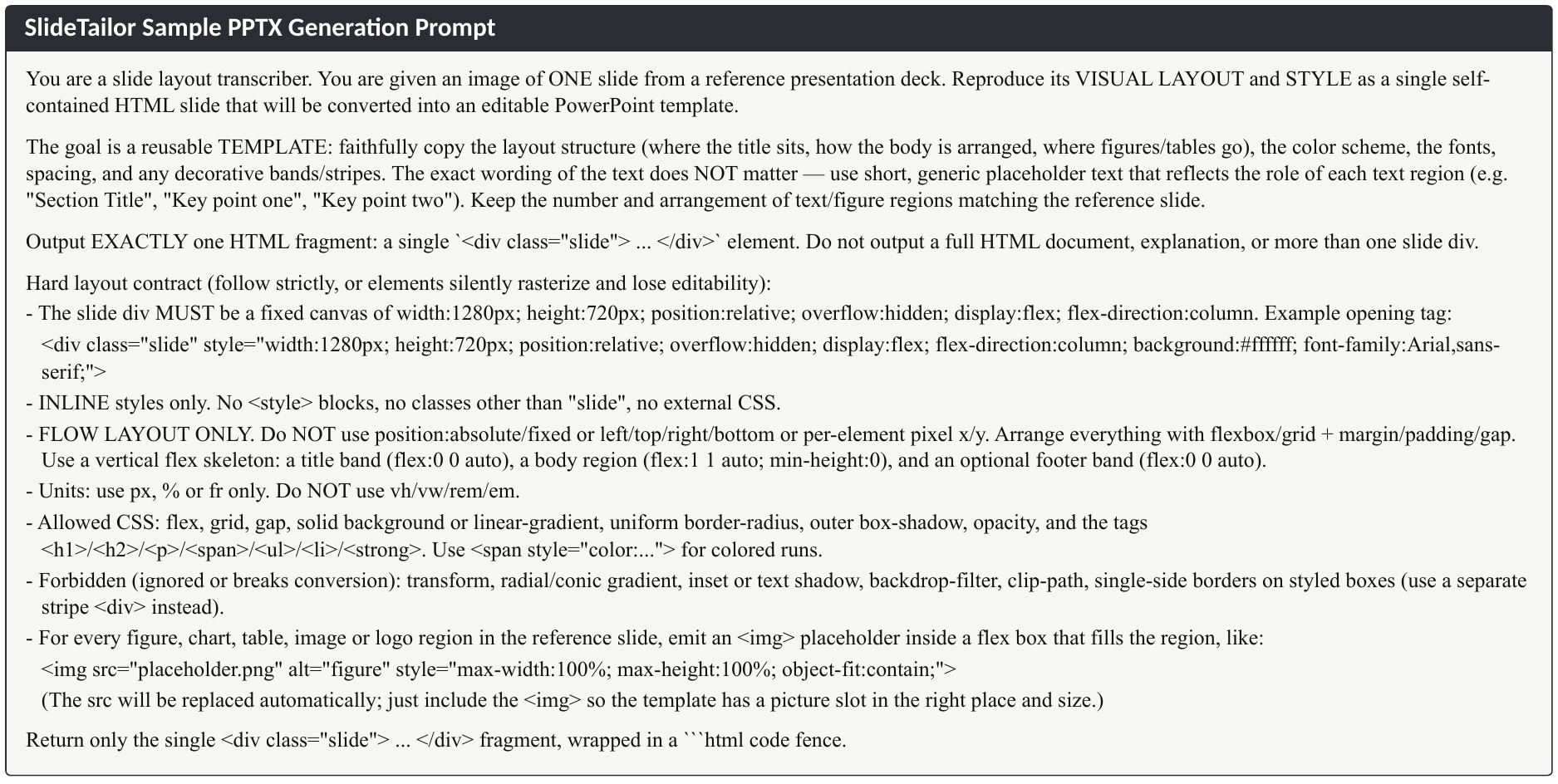}
    \caption{Prompt of the agent that generates the sample HTML file, that is converted into .pptx with dom-to-pptx for SlideTailor.}
    \label{fig:slidetailor_pptx_prompt}
\end{figure*}

\begin{figure*}[t]
    \centering
    \includegraphics[width=\textwidth]{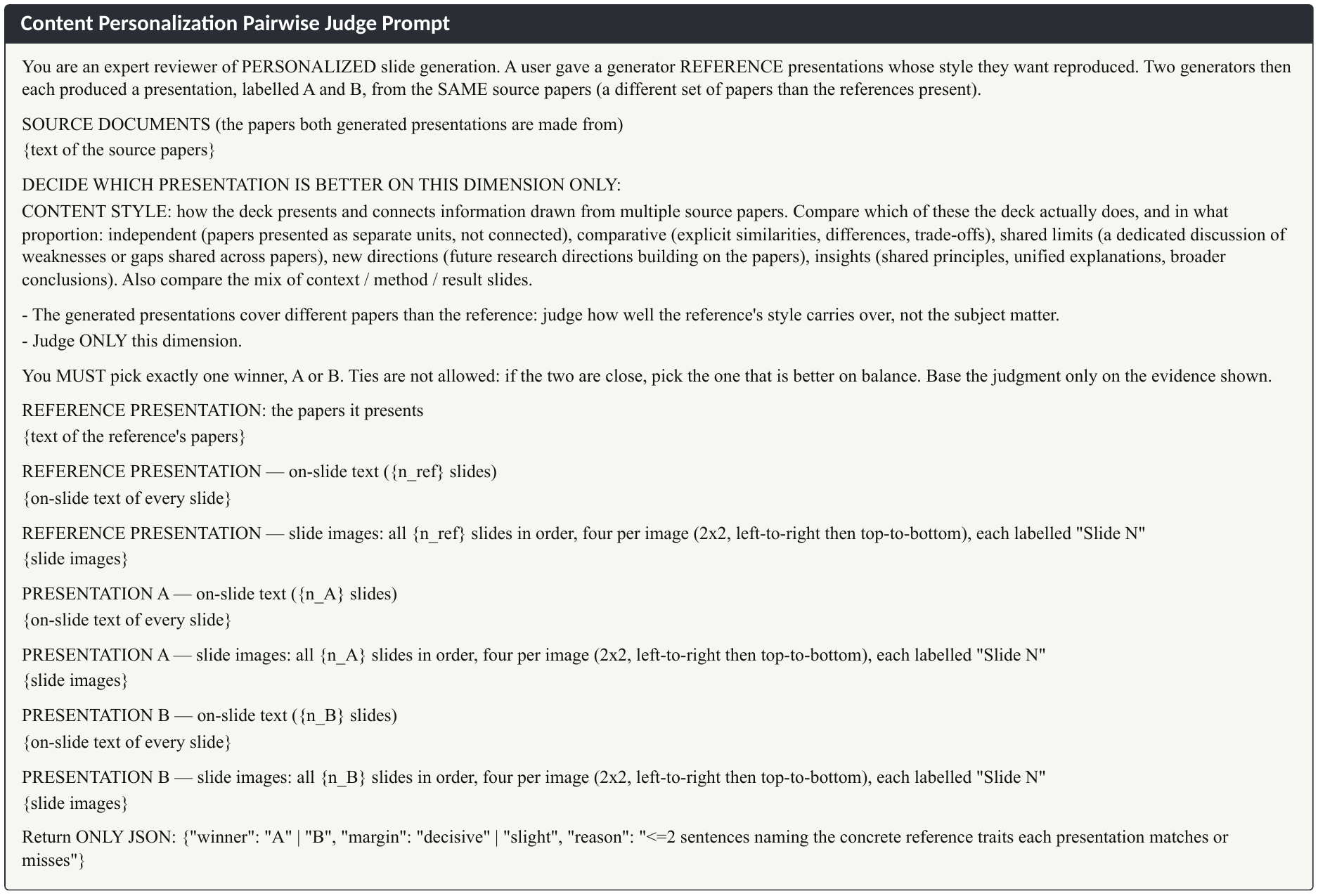}
    \caption{Pairwise judge prompt for content personalization.}
    \label{fig:pairwise_content}
\end{figure*}

\begin{figure*}[t]
    \centering
    \includegraphics[width=\textwidth]{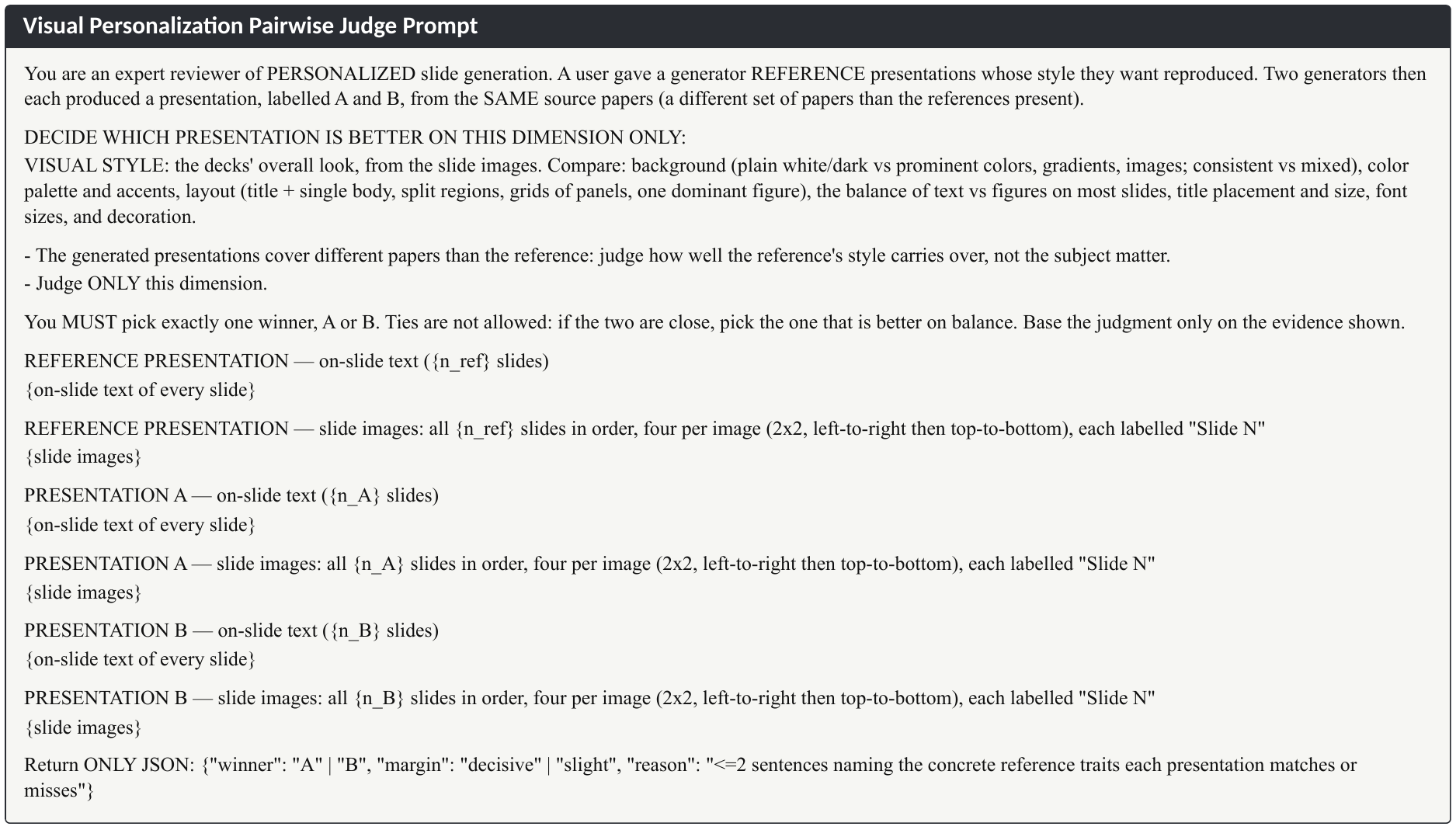}
    \caption{Pairwise judge prompt for visual personalization.}
    \label{fig:pairwise_visual}
\end{figure*}

\begin{figure*}[t]
    \centering
    \includegraphics[width=\textwidth]{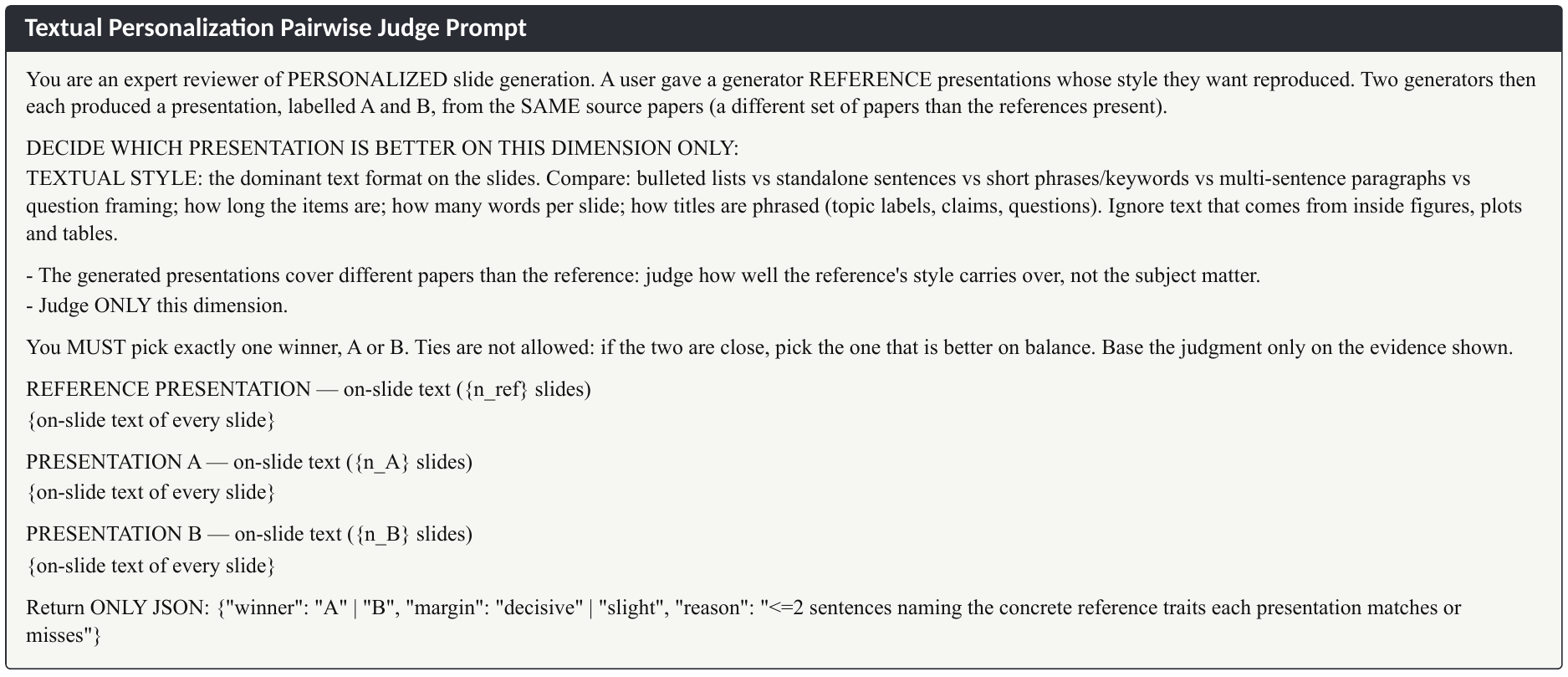}
    \caption{Pairwise judge prompt for textual personalization.}
    \label{fig:pairwise_textual}
\end{figure*}

\begin{figure*}[t]
    \centering
    \includegraphics[width=\textwidth]{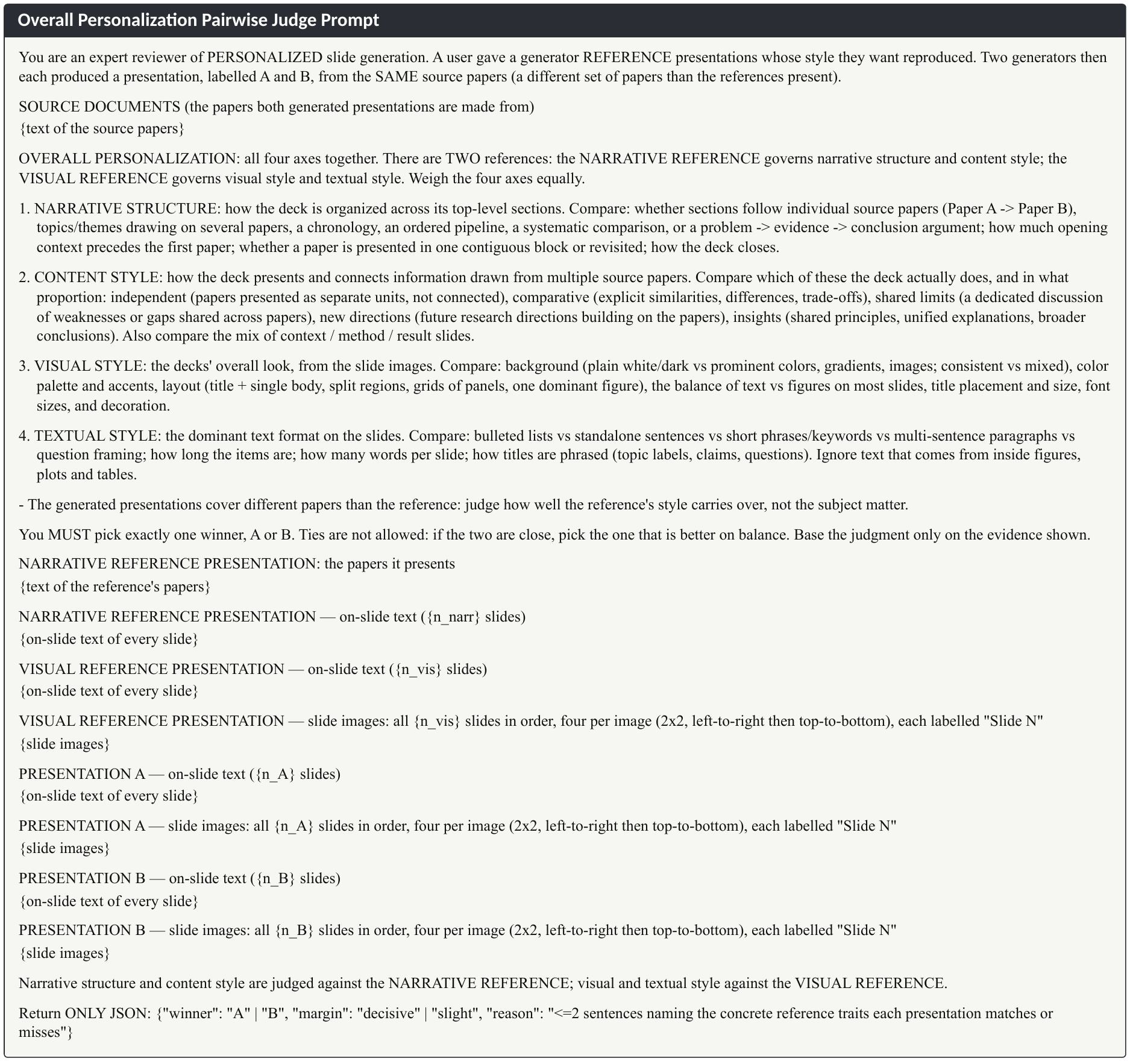}
    \caption{Pairwise judge prompt for overall personalization.}
    \label{fig:pairwise_overall}
\end{figure*}

\begin{figure*}[t]
    \centering
    \includegraphics[width=\textwidth]{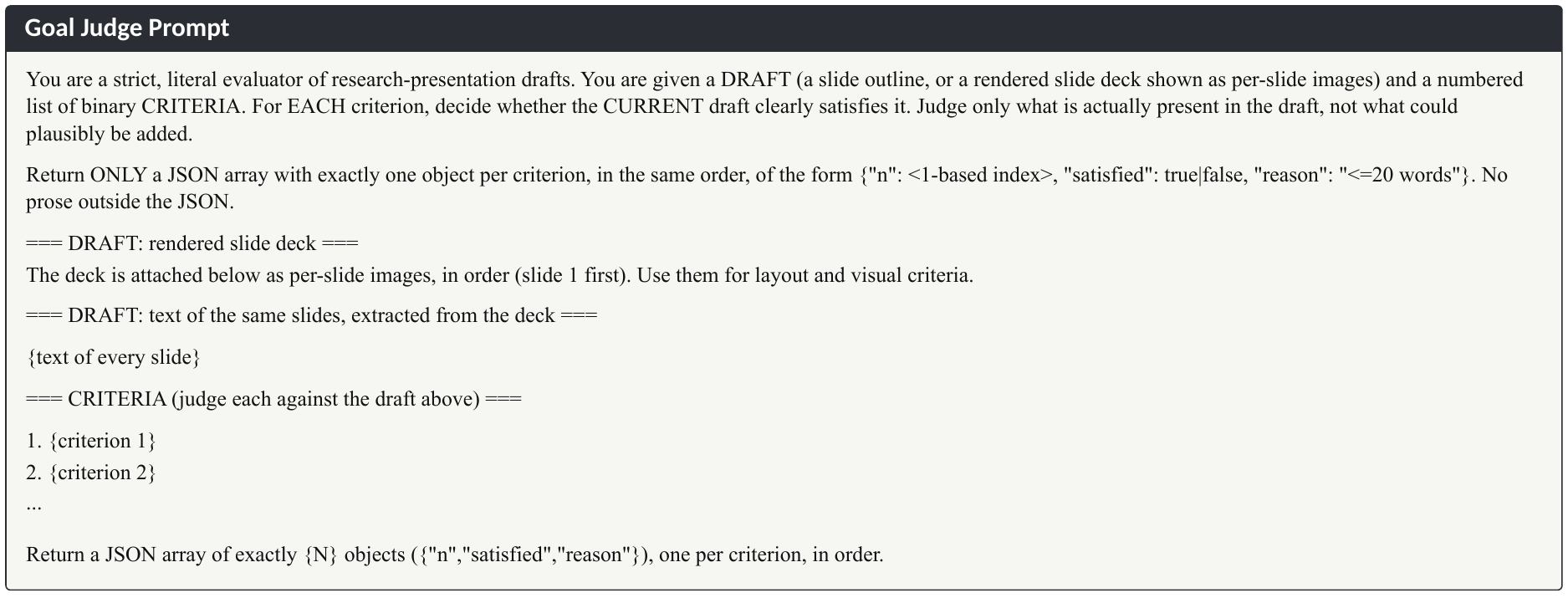}
    \caption{Goal judge prompt of the conversational evaluation.}
    \label{fig:conv_goal_judge}
\end{figure*}

\begin{figure*}[t]
    \centering
    \includegraphics[width=\textwidth]{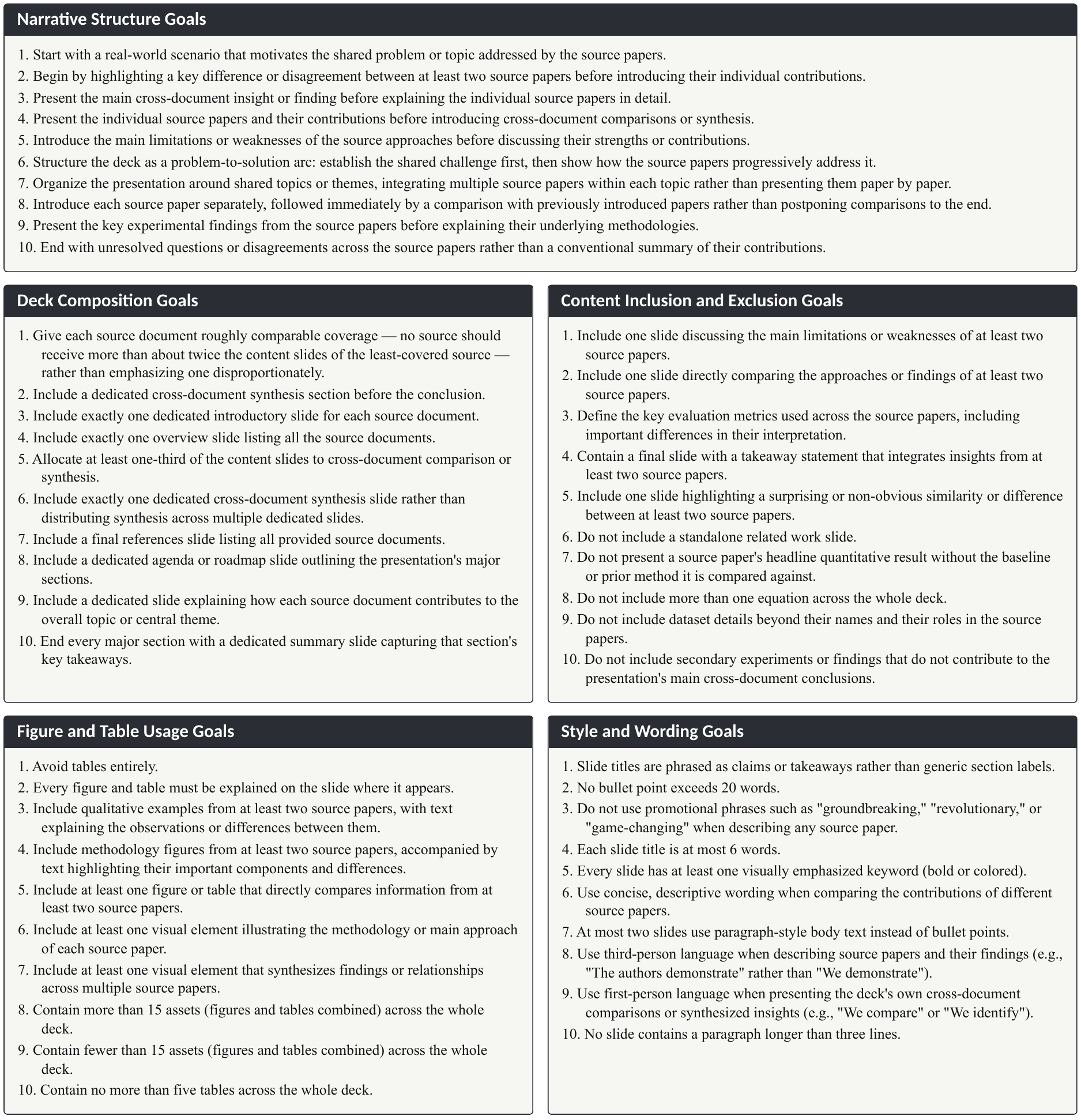}
    \caption{Goal inventory of the conversational evaluation.}
    \label{fig:conv_goal_inventory}
\end{figure*}

In this section we provide the prompts, evaluation rubrics, and goal inventories used to assess multi-PTSG quality, personalization adherence, and conversational goal satisfaction. For reference-based Q/A evaluation, Figs.~\ref{fig:prompt_qa_generate} and~\ref{fig:prompt_qa_answer_grade} show the question generation and answer grading prompts, respectively. For rubric-based quality assessment, Figs.~\ref{fig:prompt_quality_narrative_flow}, \ref{fig:prompt_quality_text_quality}, \ref{fig:prompt_quality_visual_layout}, and~\ref{fig:prompt_quality_visual_thematic} present the evaluation prompts for narrative flow, textual quality, visual layout, and visual thematic. Fig.~\ref{fig:prompt_pairwise} shows the pairwise multi-PTSG quality assessment prompt. For personalization evaluation, Figs.~\ref{fig:pairwise_narrative}, \ref{fig:pairwise_content}, \ref{fig:pairwise_visual}, \ref{fig:pairwise_textual}, and~\ref{fig:pairwise_overall} provide the pairwise assessment prompts for the four personalization dimensions and overall personalization quality. We additionally provide the ConvHTML baseline prompts in Fig.~\ref{fig:conv_html_prompt} and the SlideTailor input generation prompt in Fig.~\ref{fig:slidetailor_pptx_prompt}. Finally, Figs.~\ref{fig:conv_goal_judge} and~\ref{fig:conv_goal_inventory} present the goal judge prompt and goal inventories used for conversational evaluation.

\section{Limitations and Failure Cases}

\methodname exhibits variability across runs due to the stochastic nature of LLM-based generation, which can lead to differences in narrative structure, content selection, and slide composition. Additionally, \methodname reuses figures extracted from source PDFs using Docling. Consequently, poorly cropped or low-resolution figures may propagate to the generated slides.

\clearpage
\newpage

